\documentclass{article}

\usepackage{microtype}
\usepackage{graphicx}
\usepackage{subfigure}
\usepackage{booktabs} % for professional tables

\usepackage{hyperref}
\usepackage[accepted]{icml2023}

\usepackage{amsmath}
\usepackage{amssymb}
\usepackage{mathtools}
\usepackage{amsthm}

\usepackage[capitalize,noabbrev]{cleveref}

\theoremstyle{plain}
\newtheorem{theorem}{Theorem}
\newtheorem{proposition}{Proposition}
\newtheorem{lemma}{Lemma}

\theoremstyle{definition}
\newtheorem{definition}{Definition}
\newtheorem{assumption}{Assumption}
\theoremstyle{remark}
\newtheorem{remark}{Remark}

\newcommand{\icmlworklocation} {\textsuperscript{\dag}Work done when
interning at Huawei Cloud.}
\usepackage[textsize=tiny]{todonotes}

\icmltitlerunning{SDDM: Score-Decomposed Diffusion Models on Manifolds for Unpaired Image-to-Image Translation}

\begin{document}

\twocolumn[
% \icmltitle{Rethinking Image Conditional Score-Based Diffusion Models in The Perspective \\
% of Manifold Optimization and Multi-Objective Optimization}
\icmltitle{SDDM: Score-Decomposed Diffusion Models on Manifolds\\for Unpaired Image-to-Image Translation}

% It is OKAY to include author information, even for blind
% submissions: the style file will automatically remove it for you
% unless you've provided the [accepted] option to the icml2022
% package.

% List of affiliations: The first argument should be a (short)
% identifier you will use later to specify author affiliations
% Academic affiliations should list Department, University, City, Region, Country
% Industry affiliations should list Company, City, Region, Country

% You can specify symbols, otherwise they are numbered in order.
% Ideally, you should not use this facility. Affiliations will be numbered
% in order of appearance and this is the preferred way.
\icmlsetsymbol{star}{\dag}

\begin{icmlauthorlist}
\icmlauthor{Shikun Sun}{tsinghua,star}
\icmlauthor{Longhui Wei}{comp}
\icmlauthor{Junliang Xing}{tsinghua}
\icmlauthor{Jia Jia}{tsinghua,bnrc}
\icmlauthor{Qi Tian}{comp}
%\icmlauthor{}{sch}
%\icmlauthor{}{sch}
\end{icmlauthorlist}

\icmlaffiliation{tsinghua}{Department of Computer Science and Technology, Tsinghua University, Beijing, China}

\icmlaffiliation{comp}{Huawei Cloud.}

\icmlaffiliation{bnrc}{Beijing National Research Center for Information Science and Technology}

\icmlcorrespondingauthor{Jia Jia}{jiajia@tsinghua.edu.cn}
% \icmlcorrespondingauthor{Firstname2 Lastname2}{first2.last2@www.uk}

% You may provide any keywords that you
% find helpful for describing your paper; these are used to populate
% the "keywords" metadata in the PDF but will not be shown in the document
\icmlkeywords{}

\vskip 0.3in
]

% this must go after the closing bracket ] following \twocolumn[ ...

% This command actually creates the footnote in the first column
% listing the affiliations and the copyright notice.
% The command takes one argument, which is text to display at the start of the footnote.
% The \icmlEqualContribution command is standard text for equal contribution.
% Remove it (just {}) if you do not need this facility.

\printAffiliationsAndNotice{\icmlworklocation}  % leave blank if no need to mention equal contribution
% \printAffiliationsAndNotice{\icmlEqualContribution} % otherwise use the standard text.

\begin{abstract}
Recent score-based diffusion models (SBDMs) show promising results in unpaired image-to-image translation (I2I). However, existing methods, either energy-based or statistically-based, provide no explicit form of the interfered intermediate generative distributions. This work presents a new score-decomposed diffusion model (SDDM) on manifolds to explicitly optimize the tangled distributions during image generation. 
SDDM derives manifolds to make the distributions of adjacent time steps separable and decompose the score function or energy guidance into an image ``denoising" part and a content ``refinement" part. To refine the image in the same noise level, we equalize the refinement parts of the score function and energy guidance, which permits multi-objective optimization on the manifold. We also leverage the block adaptive instance normalization module to construct manifolds with lower dimensions but still concentrated with the perturbed reference image. SDDM outperforms existing SBDM-based methods with much fewer diffusion steps on several I2I benchmarks.

\end{abstract}

\section{Introduction}
\label{intro}
Score-based diffusion models~\cite{song2019generative,song2020score,Ho2020DDPM,nichol2021improved,bao2022analytic,lu2022dpm} (SBDMs) have recently made significant progress in a series of conditional image generation tasks. In particular, in the unpaired image-to-image translation (I2I) task~\cite{pang2021image}, recent studies have shown that a pre-trained SBDM on the target image domain with energy guidance~\cite{zhao2022egsde} or statistical~\cite{choi2021ilvr} guidance outperforms generative adversarial network~\cite{goodfellow2014generative}(GAN)-based methods~\cite{fu2019geometry, zhu2017unpaired, yi2017dualgan, park2020contrastive, benaim2017one, zheng2021spatially, shen2019towards, huang2018multimodal, jiang2020tsit, lee2018diverse} and achieves the state-of-the-art performance.

SBDMs provide a diffusion model to guide how image-shaped data from a Gauss distribution is iterated step by step into an image of the target domain. In each step, SBDM gives score guidance which, from an engineering perspective, can be mixed with energy and statistical guidance to control the generation process. However, firstly due to the stochastic differential equations of the inverse diffusion process, the coefficient of the score guidance is not changeable. Secondly how energy guidances affect the intermediate distributions is still not clear. As a result, the I2I result is often unsatisfactory, especially when iterations are inadequate. Moreover, there has yet to be a method to ensure that the intermediate distributions are not negatively interfered with during the above guidance process.

%leverage: n. 杠杆作用；优势，力量； 影响力；举债经营。v. 举债经营；借贷收购；利用。
%         在科技论文中做动词使用时主要含义是（巧妙地）借助另一种框架/模型/方法/来高效解决某一种问题。好词，但不要滥用。

%如果可以，下面一段还可以再精炼
To overcome these limitations, we propose to decompose the score function from a new manifold optimization perspective, thus better exerting the energy and statistical guidance. To this end, we present SDDM, a new score-decomposed diffusion model on manifolds to explicitly optimize the tangled distributions during the conditional image generation process. When generating an image from score guidance, an SBDM actually performs two distinct tasks, one is image ``denoising", and the other is content ``refinement'' to bring the image-shaped data closer to the target domain distribution with the same noise level. Based on this new perspective, SDDM decomposes the score function into two different parts, one for image denoising and the other for content refinement. To realize this decomposition, we take statistical guidance as the manifold restriction to get an explicit division between the data distributions in neighboring time steps. We find that the tangent space of the manifold naturally separates the denoising part and the refinement part of the score function. In addition, the tangent space can also split out the denoising part of the energy guidance, thus achieving a more explanatory conditional generation. 

Within the decomposed score functions, the content refinement part of the score function and energy functions are on an equal footing. Therefore we can treat the optimization on the manifold as a multi-objection optimization, thus avoiding the negative interference of other guidance on score guidance. To realize the score-decomposed diffusion model, we leverage the block adaptive instance normalization (BAdaIN) module to play the restriction function on the manifold, which is a stronger constraint than the widely used low-pass filter~\cite{choi2021ilvr}. With our carefully designed BAdaIN, the tangent space of the manifold provides a better division for the score and energy guidance. We also prove that our manifolds are equivalently concentrated with the perturbed reference image compared with those in~\cite{choi2021ilvr}.

To summarize, this work makes the following three main contributions:
\vspace{-4mm}
\begin{itemize}
\setlength{\itemsep}{0pt}
\setlength{\parsep}{0pt}
\setlength{\parskip}{3pt}
  \item We present a new score-decomposed diffusion model on manifolds to explicitly optimize the tangled distributions during the conditional image generation process.
  \item We introduce a multi-objective optimization algorithm into the conditional generation of SBDMs, which permits not only many powerful gradient combination algorithms but also adjustment of the score factor.
  \item We design a BAdaIN module to construct a lower dimensional manifold compared with the low-pass filter and thus provide a concrete model implementation. 
\end{itemize}
\vspace{-3mm}
With the above contributions, we have obtained a high-performance conditional image generation model. Extensive experimental evaluations and analyses on two I2I benchmarks demonstrate the superior performance of the proposed model. Compared to other SBDM-based methods, SDDM generates better results with much fewer diffusion steps.

\section{Background}
\label{background}
% This section will introduce background knowledge about SBDMs~\cite{song2020score, Ho2020DDPM, nichol2021improved, sohl2015deep}, and recent methods in the application of unpaired image-to-image translation without retraining the SBDMs. 
% % We will also introduce some fundamental conceptions in manifold optimization and multi-objective optimization.

\subsection{Score-Based Diffusion Models (SBDMs)}
\label{sbdms}
%在公式给出之前，尽量先要介绍所有使用的变量；Forward diffusion process的SDE过程难以理解的一点是没有在公式上反映出t到t+1的过程 
%那个d x就是 x_{t+1} - x_t, 如果离散化的话, 感觉除非公式去写VP-SDE的离散形式, 不然很难表示...
SBDMs~\cite{song2020score,Ho2020DDPM,dhariwal2021diffusion,zhao2022egsde} first progressively perturb the training data via a forward diffusion process and then learn to reverse this process to form a generative model of the unknown data distribution. Denoting $q(\mathbf{x}_0)$ the training set with i.i.d. samples on $\mathbb{R}^{d}$ and $q(\mathbf{x}_t)$ the intermediate distribution at time $t$, the forward diffusion process $\{\mathbf{x}_t\}_{t\in[0,T]}$ follows the stochastic differential equation (SDE):
\begin{equation}
\label{eqn, forward function}
\mathrm{d}\mathbf{x}=\mathbf{f}(\mathbf{x}, t) \mathrm{d}t+g(t) \mathrm{d}\mathbf{w},
\end{equation}
where $\mathbf{f}(\cdot, t): \mathbb{R}^{d} \rightarrow \mathbb{R}^{d}$ is the drift coefficient, $\mathrm{d}t$ denotes an infinitesimal positive timestep, $g(t)\in \mathbb{R}$ is the diffusion coefficient, and $\mathbf{w} \sim \mathcal{N}(\mathbf{0}, t\mathbf{I}_d)$ is a standard Wiener process. Denote $q_{t|0}(\mathbf{x}_t|\mathbf{x}_0)$ the transition kernel from time $0$ to $t$, which is decided by $\mathbf{f}(\mathbf{x},t)$ and $g(t)$. In practice, $\mathbf{f}(\mathbf{x},t)$ is usually an affine transformation \textit{w.r.t.} $\mathbf{x}$ so that the $q_{t|0}(\mathbf{x}_t|\mathbf{x}_0)$ is a linear Gaussian distribution and $\mathbf{x}_t$ can be sampled in one step~\cite{zhao2022egsde}.
In practice, the following VP-SDE is mostly used:
\begin{equation}
\mathrm{d} \mathbf{x}=-\frac{1}{2} \beta(t) \mathbf{x} \mathrm{d} t+\sqrt{\beta(t)} \mathrm{d} \mathbf{w},
\end{equation}
and DDPM~\cite{Ho2020DDPM,dhariwal2021diffusion} use the following discrete form of the above SDE:
\begin{equation}
\mathbf{x}_i=\sqrt{1-\beta_i} \mathbf{x}_{i-1}+\sqrt{\beta_i} \mathbf{z}_{i-1}, \quad i=1, \cdots, N.
\end{equation}

%同理，Reverse diffusion process的SDE过程难以理解的一点也是没有在公式上反映出t到t-1的过程；注意cite、citet、citep、citeauthor、citeyear、citetext、cite*等的区别
Normally an SDE is not time-reversible because the forward process loses information on the initial data distribution and converges to a terminal state distribution $q_T(\mathbf{x}_T)$. However, ~\citet{song2020score} find that the reverse process satisfies the following reverse-time SDE:
\begin{equation}
\mathrm{d} \mathbf{x}=\left[\mathbf{f}(\mathbf{x}, t)-g(t)^2 \nabla_{\mathbf{x}} \log q_t(\mathbf{x})\right] \mathrm{d} t+g(t) \mathrm{d} \overline{\mathbf{w}},
\end{equation}
where $\mathrm{d}t$ is an infinitesimal negative timestep and $\overline{\mathbf{w}}$ is a reverse-time standard Wiener process.
~\citet{song2020score} adopt a score-based model $\mathbf{s}(\mathbf{x}, t)$ to approximate $\nabla_{\mathbf{x}} \log q_t(\mathbf{x})$, \textit{i.e.} $\mathbf{s}(\mathbf{x},t) \dot= \nabla_{\mathbf{x}}\log q_t(\mathbf{x})$, obtaining the following reverse-time SDE:
\begin{equation}
\mathrm{d} \mathbf{x}=\left[\mathbf{f}(\mathbf{x}, t)-g(t)^2 \mathbf{s}(\mathbf{x}, t)\right] \mathrm{d} t+g(t) \mathrm{d} \overline{\mathbf{w}}.
\end{equation}
In VP-SDE, $q_T(\mathbf{x}_T)$ is also a standard Gaussian distribution.

For a controllable generation, it is convenient to add some guidance function $\varepsilon(\mathbf{x},t)$ to the score function and then get a new time-reverse SDE:
\begin{equation}
\!\!\!\!\mathrm{d}\mathbf{x}\!\!=\!\!\left[\mathbf{f}(\mathbf{x}, t)\!-\!g(t)^2\!\left(\mathbf{s}\left(\mathbf{x},t\right)\!+\!\nabla_{\mathbf{x}}\varepsilon\left(\mathbf{x},t\right)\right)\right]\!\mathrm{d}t+g(t)\mathrm{d}\overline{\mathbf{w}}.
\end{equation}

\subsection{SBDMs in Unpaired Image to Image Translation}
\label{subsection: other methods}
% 这里放图
% There are two main methods to guide a pretrained SBDM in the target domain in the unpaired image-to-image translation task.
Unpaired I2I aims to transfer an image from a source domain $\mathcal{Y}\subset\mathbb{R}^{d}$ to a different target domain $\mathcal{X}\subset\mathbb{R}^{d}$ as the training data. This translation process can be achieved by designing a distribution $p(\mathbf{x}_0|\mathbf{y}_0)$ on the target domain $\mathcal{X}$ conditioned on an image $\mathbf{y}_0\in\mathcal{Y}$ to transfer.

In ILVR~\cite{choi2021ilvr}, given a reference image $\mathbf{y}_0$, they refine $\mathbf{x}_t$ after each denoising step with a low-pass filter $\mathbf{\Phi}$ for the faithfulness to the reference image:
\begin{equation}
\mathbf{x}_t' = \mathbf{x}_t - \mathbf{\Phi}(\mathbf{x}_t) + \mathbf{\Phi}(\mathbf{y}_t), \mathbf{y}_t\thicksim q_{t \mid 0}(\mathbf{y}_t \mid \mathbf{y}_0).
\label{equ: ILVR low pass}
\end{equation}

In EGSDE~\cite{zhao2022egsde}, they carefully designed two energy-based guidance functions and follow the conditional generation method in~\citet{song2020score}:
\begin{equation}
\label{Eqn.original.guidance}
\begin{aligned}
\!\!\mathrm{d}\mathbf{x}\!\!=\!\!\left[\mathbf{f}(\mathbf{x}, t)\!-\!g(t)^2\!\left(\mathbf{s}(\mathbf{x},t)\!-\!\nabla_{\mathbf{x}} \varepsilon(\mathbf{x},\mathbf{y}_0,t)\right)\right]\!\mathrm{d}t\!+\!g(t)\mathrm{d}\overline{\mathbf{w}}.
\end{aligned}
\end{equation}
% EGSDE also provides an explanation as a product of experts. 
%

%------------------------------------------------------------------------
\begin{figure*}[htbp] %H为当前位置，!htb为忽略美学标准，htbp为浮动图形
 \centering %图片居中
 \includegraphics[width=0.85\textwidth]{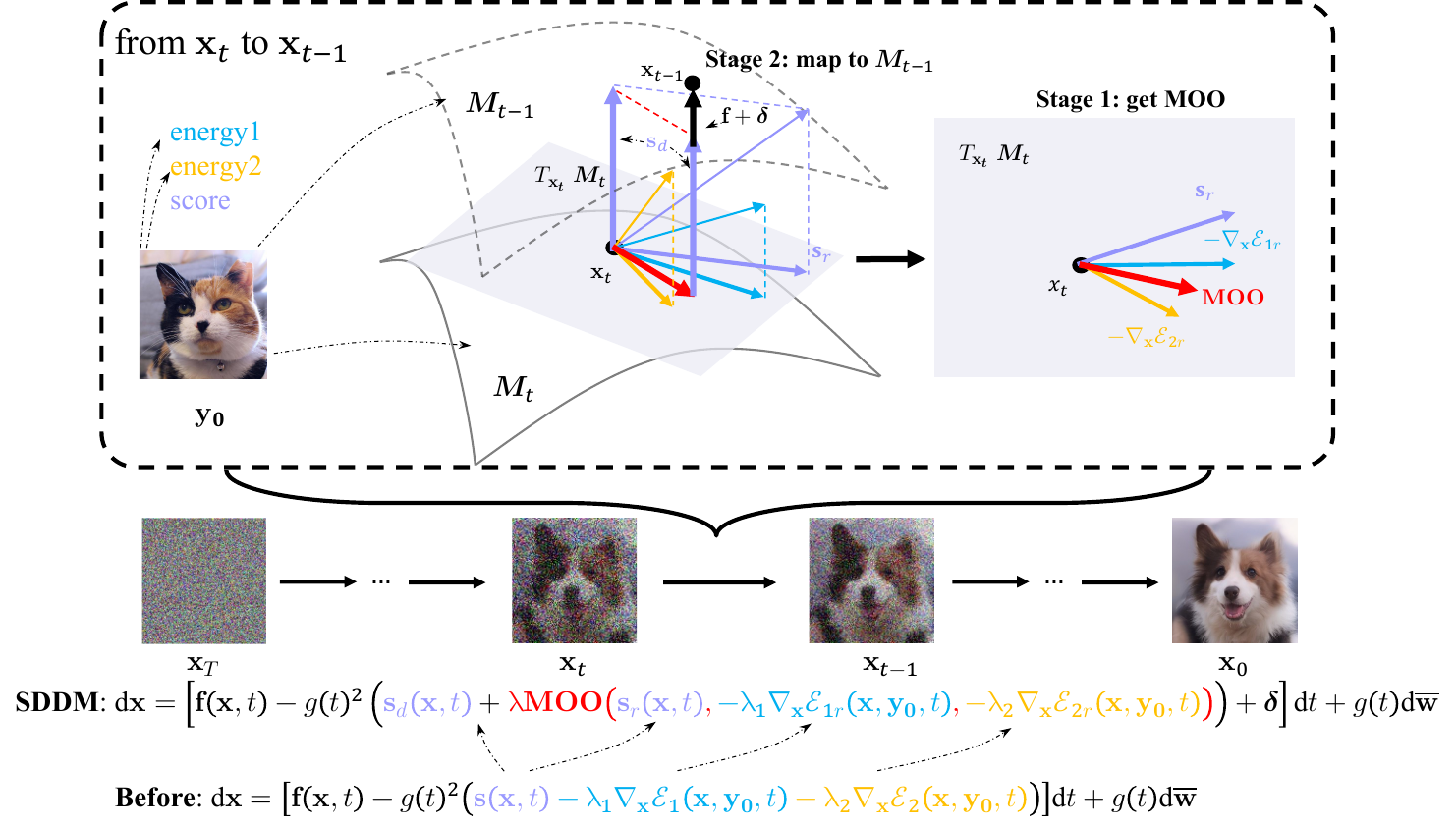} %插入图片，[]中设置图片大小，{}中是图片文件名
 \caption{The overview of our SDDM. At each time step, compared with directly adding energy guidance to the score function, we firstly use the moments of the distribution $\mathbf{y}_t$ as constraints to get the manifolds $\mathcal{M}_t$ at each time step $t$. Then, we restrict potential energy of the score function $\mathbf{s}(\mathbf{x},t)$ and energy function $\mathbf{\varepsilon}_i$ on the manifold $\mathcal{M}_t$ at $\mathbf{x}_t$ to get the components of corresponding gradients $\mathbf{s}_r(\mathbf{x},t)$ and $\nabla_{\mathbf{x_t}}\mathbf{\varepsilon}_{ir}(\mathbf{x}_t, \mathbf{y}_0, t)$ on the tangent space $T_{\mathbf{x}_t}\mathcal{M}_t$, and they are the ``refinement'' parts. Then we use multi-objective optimization viewpoint to get $\mathrm{MOO}$, the optimal sum on the tangent space near $\mathbf{x}_t$ , Finally, we restrict $\mathbf{f}(\mathbf{x}, t), \mathbf{s}(\mathbf{x},t)$, and the noise on the $N_{\mathbf{x}_t}\mathcal{M}$ to get the components pointing to the next manifold $\mathcal{M}_{t-1}$. For clarity $g(t) \mathrm{d}\overline{\mathbf{w}}$ does not appear and the restriction on $\mathcal{M}_{t-1}$ is indicated as $\boldsymbol{\delta}$.} %最终文档中希望显示的图片标题
 \label{Fig.model} %用于文内引用的标签
 \end{figure*}
 %------------------------------------------------------------------------

Notably, energy-based methods do not avoid the intermediate distribution being overly or negatively disturbed, and they both do not fully make use of the statistics of the reference image; thus the generation results may be suboptimal.

\section{Score-Decomposed Diffusion Model}
\label{sddm}
% $$
% \mathrm{d} \mathbf{x}=\left[\mathbf{f}(\mathbf{x}, t)-g(t)^2\left(s(\mathbf{x}, t)-\lambda_s \nabla_x \mathcal{E}_s\left(\mathbf{x}, \mathbf{y}_0, t\right)-\lambda_i \nabla_x \mathcal{E}_i\left(\mathbf{x}, \mathbf{y}_0, t\right)\right)\right] \mathrm{d} t+g(t) \mathrm{d} \overline{\mathbf{w}}
% $$
This section starts the elaboration of the proposed model from Eqn.~\eqref{Eqn.original.guidance}. For the choice of the guidance function \(\varepsilon(\mathbf{x},\mathbf{y}_0,t)\) in Eqn.~\eqref{Eqn.original.guidance}, we set it to the following widely adopted form~\cite{zhao2022egsde,bao2022equivariant}:
\begin{equation}
\label{Eqn.energy.guidance}
\varepsilon(\mathbf{x},\mathbf{y}_0,t)=\lambda_1\varepsilon_1(\mathbf{x},\mathbf{y}_0,t)+\lambda_2\varepsilon_2(\mathbf{x},\mathbf{y}_0,t),
\end{equation}
where \(\varepsilon_1(\cdot,\cdot,\cdot)\) and \(\varepsilon_2(\cdot,\cdot,\cdot)\) denote two different energy guidance; \(\lambda_1\) and \(\lambda_2\) are two weighting coefficients.

\subsection{Model Overview}
\label{ssec.overview}
Figure~\ref{Fig.model} overviews the main process of the proposed SDDM model. The second equation at the bottom is the equivalent SDE formulation from Eqns.~\eqref{Eqn.original.guidance} and~\eqref{Eqn.energy.guidance}. Starting with this equation, we have the first SDE in Figure~\ref{Fig.model} to indicate such a generation process. The illustration explains the two-stage optimization at time step $t$.

To explicitly optimize the tangled distributions during image generation, we use moments of the perturbed reference image $\mathbf{y}_0$ as constraints for constructing separable manifolds, thus disentangling the distributions of adjacent time steps. As shown in Figure~\ref{Fig.model}, the manifolds of adjacent time steps $t$ and $t-1, $ $\mathcal{M}_t$ and $\mathcal{M}_{t-1}$ are separable, which indicates the conditional distributions of adjacent time steps $\mathbf{x}_t$ and $\mathbf{x}_{t-1}$ are also separable. Furthermore, at time step $t$, the manifold $\mathcal{M}_t$ decompose the score function $\mathbf{s}(\mathbf{x},t)$ into the content refinement part $\mathbf{s}_r(\mathbf{x},t)$ and the image denoising part $\mathbf{s}_d(\mathbf{x}, t)$, and also separate out the content refinement parts $\nabla_{\mathbf{x}} \varepsilon_{1r}(\mathbf{x},\mathbf{y}_0,t), \nabla_{\mathbf{x}} \varepsilon_{2r}(\mathbf{x},\mathbf{y}_0,t)$ of $\nabla_{\mathbf{x}} \varepsilon_{1}(\mathbf{x},\mathbf{y}_0,t), \nabla_{\mathbf{x}} \varepsilon_{2}(\mathbf{x},\mathbf{y}_0,t)$ on the tangent space $T_{\mathbf{x}_t}\mathcal{M}_t$. Therefore, The entire optimization process at each time step is divided into two stages: one is to optimize on the manifold $\mathcal{M}_t$, and the other stage is to map to the next manifold $\mathcal{M}_{t-1}$ properly. 

In the first stage, we optimize on the manifold $\mathcal{M}_t$. We apply a multi-objective optimization algorithm to get the red vector MOO, which is the optimal direction considering the score function and energy guidance on the tangent space $T_{\mathbf{x}_t}\mathcal{M}_t$.
Then at the second stage, we use the rest of the first equation in Figure~\ref{Fig.model}, which contains $\left[\mathbf{f}(\mathbf{x}, t)-g(t)^2\mathbf{s}_d(\mathbf{x}, t)+\boldsymbol{\delta}\right] \mathrm{d} t+g(t) \mathrm{d} \overline{\mathbf{w}}$ to map the $\mathbf{x}_t + \mathrm{MOO}$ to the next manifold $\mathcal{M}_{t-1}$ properly. Note that here we use $\boldsymbol{\delta}$ to indicate the restriction on $\mathcal{M}_{t-1}$ for the consistency of form.

\subsection{Decomposition of the Score and Energy Guidance}
%%%
% Intorduce how we do the Decomposition. Give the definitions of $s_r$ and $s_d$, and some mild assumptions. Also claim the soundness of the definition via the mean of many gauss variable has small Var.
%%%

% \begin{proposition}
% Let $\mathbf{x} = (x_1, x_2, ..., x_n)^T \sim \mathcal{N}(\mathbf{0}, \mathbf{I})$, $\mathbf{N}\left(\mathbf{x}\right)$ is the function to normalize $\mathbf{x}$'s mean or variance to the corresponding first order or second order moment and $\mathbf{N}_i\left(\mathbf{x}\right)$ is the $i$th component of $\mathbf{N}\left(\mathbf{x}\right)$, then we have: \\
% $\forall \varepsilon, \xi > 0, \exists N, \forall n > N$,\\
% $ \mathbf{P}\left[| \mathbf{N}_i \left(\mathbf{x}\right) - x_i| < \varepsilon \right] > 1 - \xi, 1\le i \le n$.
% \end{proposition}

%Some basic definitions of manifolds are shown in Appendix~\ref{app: BK about manifolds}. More details can be found in~\cite{lee2010introduction, tu2011manifolds}.

Given a score function $\mathbf{s}(\mathbf{x})=\nabla_{\mathbf{x}}\log p(\mathbf{x})$ on $\mathbb{R}^d$, suppose $\mathcal{M}$ is a smooth, compact submanifold of $\mathbb{R}^d$. We let $p_{\mathcal{M}}(\mathbf{x})$ is the corresponding probability distribution restricted on $\mathcal{M}$. Then we have the following definitions:

\begin{definition} \textbf{The tangent score function $\mathbf{s}_r(\mathbf{x})$.}

  $\mathbf{s}_r(\mathbf{x}):=\nabla_{\mathbf{x}}\log p_{\mathcal{M}}(\mathbf{x})$,
\end{definition}
which is the score function on the manifold. If there is a series of manifolds $\{\mathcal{M}_t\}$, and the original score function is denoted $\mathbf{s}(\mathbf{x}, t)$, we denote $\mathbf{s}_r(\mathbf{x}, t)$ the tangent score function on $\mathcal{M}_t$.

\begin{definition} \textbf{The normal score function $\mathbf{s}_d(\mathbf{x})$.}

$\mathbf{s}_d(\mathbf{x}):=\mathbf{s}(\mathbf{x})|_{N_{\mathbf{x}}\mathcal{M}}$,
\end{definition}
which is the score function on the normal space of the manifold. We also denote $\mathbf{s}_d(\mathbf{x}, t)$ the normal score function on the manifold $\mathcal{M}_t$.

Then we have the following score function decomposition:
\begin{lemma}
  \label{lemma: decouple of score}
  $\mathbf{s}(\mathbf{x})=\mathbf{s}_r(\mathbf{x}) + \mathbf{s}_d(\mathbf{x})$,
\end{lemma}
which can be derived when knowing $\mathbf{s}_r(\mathbf{x})=\mathbf{s}(\mathbf{x})|_{T_{\mathbf{x}}\mathcal{M}}$.

Normally this division is meaningless because the manifolds of adjacent time steps are coupled with each other. Previous researchers usually treat the entire $\cup_t \mathbf{x}_t$ as an entire manifold~\cite{liu2022pseudo}, or use strong assumptions~\cite{chung2022improving}. However, in some conditional generation tasks, for example, the image-to-image transition task, a given reference image $\mathbf{y}_0$ can provide compact manifolds at different time steps, and manifolds of adjacent time steps can be well separated. In this situation, the tangent score function can be treated as a refinement part on the manifold. The normal score function is part of the mapping function between manifolds of adjacent time steps. 

We have Proposition~\ref{prop: compact_manifolds} to describe the manifolds.
\begin{proposition}
 \label{prop: compact_manifolds}
 At time step $t$, for any single reference image $\mathbf{y}_0$, the perturbed distribution $q_{t \mid 0}(\mathbf{y}_t \mid \mathbf{y}_0)$ is concentrated on a compact manifold $\mathcal{M}_t\subset \mathbb{R}^d$ and the dimension of $\mathcal{M}_t \le d-2$ when $d$ is large enough. Suppose the distributions of perturbed reference image $\mathbf{y_t} = \hat{\alpha}_t\mathbf{y}_0 + \hat{\beta}_t \mathbf{z}_t$, where $\mathbf{z}_t \sim \mathcal{N}(\mathbf{0}, \mathbf{I})$. The following statistical constraints define such (d-2)-dim $\mathcal{M}_t$.
 \begin{equation}
 \label{eqn.general.yt}
  \begin{aligned}
   \operatorname{\mu}[\mathbf{x}_t] &= \hat{\alpha}_t \mu[\mathbf{y}_0],\\
   \operatorname{Var}[\mathbf{x}_t] &= \hat{\alpha}^2_t\operatorname{Var}[\mathbf{y}_0] + \hat{\beta}^2_t.
  \end{aligned}
 \end{equation}
\end{proposition}

Proposition~\ref{prop: compact_manifolds} shows that we can use statistical constraints to define concentrated manifolds with lower dimensions than $\mathbb{R}^d$. We can also use the chunking trick to lower the manifold dimensions, which will be introduced in Section~\ref{section: chunking trick}. Therefore, we can use such manifolds to represent the maintenance of the statistics, which indicates that the tangent space $T_{\mathbf{x}_t}\mathcal{M}_t$ can separate the ``refinement'' part well.

We also have Lemma~\ref{lemma:separable} to show that perturbed distributions of adjacent time steps, $\mathbf{y}_t$ and $\mathbf{y}_{t-1}$ can be well separated.

\begin{lemma}
\label{lemma:separable}
 With the $\mathcal{M}_t$ defined in Proposition~\ref{prop: compact_manifolds}, assume $t \ne t'$, Then $\mathcal{M}_t$ and $\mathcal{M}_{t'}$ can be well separated. Rigorously, $\forall \varepsilon > 0, \exists \mathcal{M}_d$ divide the $\mathbb{R}^d$ into two disconnect spaces $\mathcal{A}, \mathcal{B}$, where $\mathcal{M}_t \in \mathcal{A}$ ,and $\mathcal{M}_{t'} \in \mathcal{B}$.
\end{lemma}

% We can then assume that $\mathcal{M}_t \subset \{\mathbf{x}_t \in \mathbb{R}^d \mid q(\mathbf{x}_t) = q_{t\mid 0}(\mathbf{x}_t\mid \mathbf{x}_0) q_0(\mathbf{x}_0)\}$, which is a natural assumption when the $\mathbf{y}_0$ is a meaningful picture. 
Therefore, we can use $\mathcal{M}_t$ to decompose $\mathbf{s}$ into $\mathbf{s}_r$ and $\mathbf{s}_d$ approximately. More generally, we can decouple the optimization space with the tangent space $T_{\mathbf{x}_t} \mathcal{M}_t$. With $T_{\mathbf{x}_t} \mathcal{M}_t$, we can operate the score function of SBDM and energy more elaborately. We can also split the ``refinement" part out, thus preventing the ``denoising" part of the score function from being overly disturbed. 
\subsection{Stage 1: Optimization on Manifold}

% 介绍利用分块化加速的算法

Firstly, we will give some main definitions about manifold optimization and muti-objective optimization in our task. We use restriction $\mathbf{R}_{\mathbf{x}_t}$ represent the function that maps the points on $T_{\mathbf{x}_t} \mathcal{M}_t$ near $\mathbf{x}_t$ to the manifold $\mathcal{M}_t$, which is normally an orthogonal projection onto $\mathcal{M}_t$. 

\begin{definition}
 \textbf{Manifold optimization.}

 Manifold optimization~\cite{hu2020brief} is a task to optimize a real-valued function $f(\mathbf{x})$ on a given Riemannian manifold $\mathcal{M}$. The optimized target is: 
 \begin{equation}
  \min _{\mathbf{x} \in \mathcal{M}} f(\mathbf{x}).
  \end{equation}
\end{definition}

Because that given $t$, the score function $\mathbf{s}(\mathbf{x} ,t)$ is an estimation of $\nabla_{\mathbf{x}} \log q_t(\mathbf{x})$, and we can use $\log q_t(\mathbf{x})$ as the potential energy of $\mathbf{s}(\mathbf{x},t)$, so are the guidance of energy funcitons. Then Stage 1 is a manifold optimization.
\begin{definition}
 \textbf{Pareto optimality on the manifold.}

 Consider $\mathbf{x}_t, \widehat{\mathbf{x}}_t \in \mathcal{M}_t$,
 \vspace{-3mm}
 \begin{itemize}
 \setlength{\itemsep}{0pt}
\setlength{\parsep}{0pt}
\setlength{\parskip}{3pt}
  \item $\mathbf{x}_t$ dominates $\widehat{\mathbf{x}}_t $ if $\mathbf{s}_r(\mathbf{x}_t , t) \ge \mathbf{s}_r(\widehat{\mathbf{x}}_t, t)$, $\mathbf{\varepsilon}_{ir}(\mathbf{x}_t, \mathbf{y}_0, t) \le \mathbf{\varepsilon}_{ir}(\widehat{\mathbf{x}}_t, \mathbf{y}_0, t)$ for all $i$, and not all equal signs hold at the same time.
  \item A solution $\mathbf{x}_t $ is called Pareto optimal if there exists no solution $\widehat{\mathbf{x}}_t$ that dominates $\mathbf{x}_t $.
 \end{itemize}
\end{definition}
\vspace{-3mm}
Then, the goal of multi-objective optimization is to find the Pareto optimal solution. The local Pareto optimality can also be reached via gradient descent like single-objective optimization. We just follow the multiple gradient descent algorithm (MGDA)~\cite{desideri2012multiple}. MGDA also leverages the Karush-Kuhn-Tucker (KKT) conditions for the multi-objective optimization, which in our task is that:

\begin{theorem}
 \textbf{K.K.T. conditions on a smooth manifold.}

 At time step $t$ on the tangent space $T_{\mathbf{x}_t}\mathcal{M}_t$, there $\exists \alpha, \beta^1, \beta^2, ..., \beta^m \ge 0$ such that $\alpha + \sum_{i = 1}^m \beta^i = 1$ and $\alpha \mathbf{s}_{r}(\mathbf{x_t}, t) = \sum_{i = 1}^m \beta^i \nabla_{\mathbf{x_t}} \varepsilon_{ir}(\mathbf{x}_t, \mathbf{y}_0, t)$ , where $\mathbf{s}_r(\mathbf{x_t}, t)$ are the fractions of $\mathbf{s}(\mathbf{x_t}, t)$ on the tangent space and $\varepsilon_{ir}(\mathbf{x}_t, \mathbf{y}_0, t)$ are functions restricted on the manifold $\mathcal{M}_t$.
\end{theorem}

All points that satisfy the above conditions are called Pareto stationary points. Every Pareto optimal point is Pareto stationary point, while the reverse is not true. \citet{desideri2012multiple} showed that the optimization solution for the problem :
\begin{equation}
 \min _{\substack{\alpha, \beta^1, \ldots, \beta^m \ge 0 \\\alpha+ \beta^1+ \ldots+ \beta^m = 1} }\left\{\left\| \alpha \mathbf{s}_r(\mathbf{x}_t, t) - \sum_{i=1}^m \beta^i \nabla_{\mathbf{x_t}} \varepsilon_{ir}(\mathbf{x}_t, \mathbf{y}_0, t)\right\|_2^2\right\}
 \label{equ: optimization target}
 \end{equation}
gives a descent direction that improves all tasks or gives a Pareto stationary point. For a balanced result, we normalize all gradients first.

However, In our task, we can search Pareto stationary points in $\mathbf{B}_{\epsilon}(\mathbf{x}_t) \cap \mathcal{M}_t$ for a small $\epsilon$ because we have many time steps of different manifolds. $\mathbf{B}_{\epsilon}(\mathbf{x}_t)$ is an open ball with center $\mathbf{x}_t$, radius $\epsilon$.
% \begin{proposition}
%  $\mathbf{MOO}(s_r, g_1, g_2, ..., g_n) \le s_r$
% \end{proposition}
% \begin{proposition}
%  If $s_r$ has an acceptable error in the restriction step, then the error of $\mathbf{MOO}(s_r, g_1, g_2, ..., g_n)$ is also acceptable.
% \end{proposition}
% The above theorem shows that the $\mathbf{MOO}$ algorithm will not overly interfere

We have the following algorithm:

\begin{algorithm}[htbp]
 \caption{Multi-Objective Optimization on Manifold}
 \label{alg:opt on manifold}
\begin{algorithmic}[1]
 \STATE {\bfseries Input:} stepsize $\lambda$, current $\mathbf{x}_t$, refinement score $s_r$, energy funcitons $\varepsilon_{ir}$ on $\mathcal{M}_t, i = 1, \ldots, m$, $\epsilon$
 \STATE {\bfseries Output:} $\mathbf{x}_t^*$
 % \FOR{$i=1$ {\bfseries to} $T$ }
 % \STATE x = 1 \COMMENT{All Noun Phrases of D}
 % \ENDFOR

 % \FUNCTION{name}
 % \STATE body
 % \ENDFUNCTION
 \STATE Initialize $\mathbf{x}_t^* = \mathbf{x}_t$
 \REPEAT
 \STATE $\nabla_{\mathbf{x}_t^*} \varepsilon_{ir}' = \lambda_i \frac{\|\mathbf{s}_r(\mathbf{x}_t^*, t) \|}{\|\nabla_{\mathbf{x}_t^*} \varepsilon_{ir}(\mathbf{x}_t^*, \mathbf{y}_0, t)\|} \nabla_{\mathbf{x}_t^*} \varepsilon_{ir}(\mathbf{x}_t^*, \mathbf{y}_0, t)$
 \STATE Get the min value $v$ of eq. \eqref{equ: optimization target} and corresponding $\alpha, \beta^1, \ldots, \beta^m$
 \IF{$v == 0$}
 \STATE \textbf{return} $\mathbf{x}_t^*$
 \ENDIF
 \STATE $\delta = \alpha \mathbf{s}_r(\mathbf{x}_t^*, t) - \sum_{i=1}^m \beta^i \nabla_{\mathbf{x}_t^*} \varepsilon_{ir}'$
 \STATE $\mathbf{x}'_t = \mathbf{x}_t^* + \lambda \delta$
 \STATE $\mathbf{x}_t^* = \mathbf{R_{\mathbf{x}_t^*}}(\mathbf{x}'_t)$
 \UNTIL{$\mathbf{x}_t^* \notin \mathbf{B}_{\epsilon}(\mathbf{x}_t)$}
 \STATE \textbf{return} $\mathbf{x}_t^*$
\end{algorithmic}
\end{algorithm}

\begin{remark}
\label{remark: simulate}
 We can use $\mathbf{f}(\mathbf{x}_t, t)$ and $\mathbf{s}_d(\mathbf{x}_t, t)$ to approximate $\mathbf{f}(\mathbf{x}_t^*, t)$ and $\mathbf{s}_d(\mathbf{x}_t^*, t)$ when $\epsilon$ is small. 
\end{remark}

\begin{remark}
Notably, EGSDE~\cite{zhao2022egsde} applies coefficients directly on the guidance vectors, and DVCE~\cite{augustin2022diffusion} uses coefficients after the normalization on the guidance vectors. We can also provide coefficients $\lambda_i$s for normalized energy vectors to change the impact of the vectors. A smaller norm means greater impact, as mentioned in~\cite{desideri2012multiple}.
\end{remark}

\subsection{Stage 2: Transformation between adjacent manifolds}
After the optimization on the manifold $\mathcal{M}_t$, we get $\mathbf{x}_t^*$ that dominates $\mathbf{x}_t$, then we use $\mathbf{f}(\mathbf{x}_t^*, t)$, the ``denoising'' part score function $\mathbf{s}_d(\mathbf{x}^*, t)$, reverse-time noise and restriction function on $\mathcal{M}_{t-1}$ to map to the adjacent manifold $\mathcal{M}_{t-1}$.

Firstly, we have the following proposition to describe the properties of the adjacent map.
\begin{proposition}
\label{proposition: properties of adjacent map}
Suppose the $\mathbf{f}(\cdot, \cdot)$ is affine. Then the adjacent map has the following properties:
 \begin{itemize}
  \item $\exists ! \mathbf{v_{\mathbf{x}_t}} \in N_{\mathbf{x}_t} \mathcal{M}_t$ that $\mathbf{x}_t + \mathbf{v_{\mathbf{x}_t}} \in \mathcal{M}_{t - 1}$.
  \item $  N_{\mathbf{x}_t} \mathcal{M}_t = N_{\mathbf{x}_t + \mathbf{v_{\mathbf{x}_t}}} \mathcal{M}_{t-1}$ .
  \item $\mathbf{v_{\mathbf{x}_t}}$ is a transition map from $ T_{\mathbf{x}_t} \mathcal{M}_t$ to $T_{\mathbf{x}_t + \mathbf{v}} \mathcal{M}_{t-1}$ .
  \item $\mathbf{v_{\mathbf{x}_t}}$ is determined with $\mathbf{f}(\cdot, \cdot), \mathbf{x_t}, \mathbf{y}_0$ and $g(\cdot)$. 
  % \item If $\|v\| >> \|\mathbf{x}_t^* - \mathbf{x}_t\|$, 
 \end{itemize} 
\end{proposition}

However, if we use $\mathbf{v_{\mathbf{x}_t}}$ as the adjacent map, we will lose the impact of $\mathbf{s}_d$ and the reverse-time noise. Therefore, we follow the reverse SDE, using the extra part of which on the normal space $N_{\mathbf{x}_t}\mathcal{M}_t$ and a restriction function on $\mathcal{M}_{t-1}$ as the adjacent map, we denote this part as $\mathbf{v_{\mathbf{x}_t}^*}$.

Finally, we have Algorithm~\ref{alg:generation_with_sddm} in the following to generate images with our proposed SDDM.

\begin{algorithm}[htbp]
 \caption{Generation with SDDM}
 \label{alg:generation_with_sddm}
\begin{algorithmic}[1]
 \STATE {\bfseries Input:} time steps $T$, milstone time step $T_0$, stepsize $\lambda$, score function $\mathbf{s}$, energy funcitons $\varepsilon_{i}, i = 1, \ldots, m$, small $\epsilon$, reference image $\mathbf{y}_0$
 \STATE {\bfseries Output:} generated image $\mathbf{x}_0$
 \STATE Initialize $\mathbf{x}_T \in \mathcal{M}_T$

 \FOR{$t=T$ {\bfseries to} $T_0$ }
 \STATE Divide the $\mathbb{R}^{d}$ into two orthogonal spaces $T_{\mathbf{x}_t}\mathcal{M}_t$ and $N_{\mathbf{x}_t}\mathcal{M}_{t}$.
 \STATE Calculate $\mathbf{s}_r(\cdot, \cdot)$ and $\varepsilon_{ir}(\cdot, \cdot, \cdot)$
 \STATE Optimize on manifold $\mathcal{M}_t$ with algorithm~\ref{alg:opt on manifold}, and get the output $\mathbf{x}_t^*$
 \STATE Apply the time-reverse SDE on the $N_{\mathbf{x}_t^*}\mathcal{M}_{t}$ and then use the restriction function $\mathbf{R}$ on manifold $\mathcal{M}_{t-1}$ to map the $\mathbf{x}_{t}^*$ to $\mathbf{x}_{t-1} \in \mathcal{M}_{t-1}$
 \ENDFOR
 \FOR{$t=T_0 -1$ {\bfseries to} $1$ }
 \STATE Apply unconditional time-reverse SDE from $\mathbf{x_t}$ to $\mathbf{x}_{t-1}$
 \ENDFOR
 \STATE \textbf{return} $\mathbf{x}_0$
 % \FUNCTION{name}
 % \STATE body
 % \ENDFUNCTION
\end{algorithmic}
\end{algorithm}

\begin{remark}
\label{remark, f is in normal space}
    If $\mathbf{f}(\mathbf{x}_t, t)$ is linear to $\mathbf{x}_t$, then $\mathbf{f}(\mathbf{x}_t, t) \in N_{\mathbf{x}_t} \mathcal{M}$.
\end{remark}

\begin{remark}
 When the $\epsilon$ is small, we can just use $N_{\mathbf{x}_t} \mathcal{M}_t$ to approximate $N_{\mathbf{x}_t^*} \mathcal{M}_t$.
\end{remark}
\begin{remark}
 Suppose $\|\mathbf{s}_r\|$ is $\mathbf{o}\left(\|\mathbf{v_{\mathbf{x}_t}^*}\|\right)$, and $\epsilon$ is $\mathbf{O}(\|\mathbf{s}_r\|)$. We can ignore the restriction step in algorithm~\ref{alg:opt on manifold}.
 \label{remark ignore normalization }
\end{remark}

\begin{remark}
 At time step $T_0$, we can set larger $\epsilon$ for better results.
\end{remark}
% Although we have such a perfect map between adjacent manifolds, we lose the information of $\mathbf{s}_d$ and the stochasticity of the $\mathrm{d}\mathbf{w}$, therefore, we use $(\mathbf{f}(\mathbf{x}_t, t) + \mathbf{s}_d)\mathrm{d} t + \mathrm{d} \mathbf{w}$ and restriction function on $\mathcal{M}_{t-1}$ as the transformation between adjacent manifolds.

% we have the following proposition:
% \begin{proposition}
%  If the curvature of the manifold is relatively small
% \end{proposition}
% Give a theorem, under the assumption $s_d >> s_r$ ( We have experiment to show this) and $s_d$ is small enough. the $s_d$ is almost an affine map between submanifolds of adjacent points of time. and then give the entire algorithm of one denoising step. need another experiment.

% Maybe: Theorem: Manifold Optimization and Multi-Objective Optimization can be combined with.

\section{Implementations}
\label{impl}
% In the following sections, we will introduce the chunking trick, how we design our manifolds in detail, and the energy function without extra training.
\textbf{Chunking Trick.}
\label{section: chunking trick}
The chunking trick is an easy but powerful trick to reduce the dimensions of the manifolds in high-dimensional space problems, like the generation of images. We will divide the image shape $C\times H \times W$ into blocks $N \times N$, and the shape will be like $CN^2 \times \frac{H}{N} \times \frac{W}{N}$, and the manifold will be the direct product of $CN^2$ manifolds at each $ \frac{H}{N} \times \frac{W}{N}$-sized block, we index them with $(c, i, j)$. This trick has the following advantages:
\begin{itemize}
 \item We can easily get the $T \mathcal{M}$ and the $N \mathcal{M}$, which are also the direct product of each block's $T \mathcal{M}$ and $N \mathcal{M}$.
 \item We can control the impact of the reference image on the generation process.
 \item We can optimize on block level and lower the impact of other distant blocks.
 % \item We can accelerate the generation process. 不是很明显
\end{itemize}

\textbf{Manifold Details.}
For each chunked $\frac{H}{N} \times \frac{W}{N}$-sized block of the image, we use the first-order and second-order moments to restrict the statistics of the pixels of the block to get a $(\frac{H}{N} \times \frac{W}{N} -2)-dim$ manifold.
In particular, we denote the $\frac{H}{N}\times \frac{H}{N}$ as $d$. Suppose $\mathbf{y}_t = \sqrt{\bar{\alpha}_t} \mathbf{y}_0+ \sqrt{1-\bar{\alpha}_t} \mathbf{z_t}$, Then, $\mathbf{y}_t \sim \mathcal{N}\left( \sqrt{\bar{\alpha}_t} \mathbf{y}_0,\left(1-\bar{\alpha}_t\right) \mathbf{I}\right)$, and the $\mathbf{y}^{c, i, j}_t \sim \mathcal{N}\left( \sqrt{\bar{\alpha}_t} \mathbf{y}_0^{c, i, j},\left(1-\bar{\alpha}_t\right) \mathbf{I}\right)$. Then, the manifold $\mathcal{M}_t^{c,i,j}$of block $(c, i, j)$ is restricted with:
\begin{equation}
 \begin{aligned}
  \operatorname{\mu}[\mathbf{x}_t^{c, i, j}] &= \sqrt{\bar{\alpha}_t}\operatorname{\mu}[\mathbf{y}_0^{c, i, j}],\\
  \operatorname{Var}[\mathbf{x}_t^{c, i, j}] &= \bar{\alpha}_t\operatorname{Var}[\mathbf{y}_0^{c, i, j}] + (1-\bar{\alpha}_t).
 \end{aligned}
 \label{equ: restriction}
\end{equation}

And the restrictions of Eqn.~\ref{equ: restriction}, $\mathcal{M}_t^{c,i,j}$ is a $(d-2)$ dimensional hypersphere.
Then we can formulate $\mathcal{M}_t$ as:
\begin{equation}
 \mathcal{M}_t = \otimes_{c,i,j}\mathcal{M}_t^{c,i,j}.
\end{equation}
The $\otimes$ denotes the direct product. \citet{huang2017arbitrary} use the AdaIN module to transfer neural features as 
\begin{equation}
 \operatorname{AdaIN}(\mathbf{x}, \mathbf{y})=\sigma(\mathbf{y})\left(\frac{\mathbf{x}-\operatorname{\mu}(\mathbf{x})}{\sigma(\mathbf{x})}\right)+\operatorname{\mu}(\mathbf{y}).
\end{equation}
Based on that, we leverage a useful module of BAdaIN as the restriction function on any $T_{\mathbf{x}_t}\mathcal{M}_t$:
\begin{equation}
\begin{aligned}
 \operatorname{BAdaIN}(\mathbf{x}_t, \mathbf{y}_t)&=\otimes_{c,i,j}\sigma(\mathbf{y}_t^{c,i,j})\left(\frac{\mathbf{x}_t^{c,i,j}-\operatorname{\mu}(\mathbf{x}_t^{c,i,j})}{\sigma(\mathbf{x}_t^{c,i,j})}\right) \\ 
 &+\otimes_{c,i,j}\operatorname{\mu}(\mathbf{y}_t^{c,i,j})
\end{aligned}
\end{equation}
In practice, we use the distribution moments of the perturbed reference image to simplify the calculation and eliminate randomness after knowing the relationship between the perturbed and original reference images, as in Eqn.~\eqref{eqn.general.yt}.

We have Lemma~\ref{lemma:detailed_compact_manifold} to describe Proposition~\ref{prop: compact_manifolds} in detail:

\begin{lemma}
\label{lemma:detailed_compact_manifold}
 $\forall \epsilon, \xi > 0, \exists D >0, \forall d > D$ we have:
 
 $$\mathcal{P}\left(d_2\left(\mathbf{y}_t^{c, i, j}, \mathcal{M}_t^{c, i, j}\right) < \epsilon \sqrt{d}\right) > 1 - \xi,$$
\end{lemma}
where $d$ is the dimension of the Euclid space $\mathbf{y}_t^{c, i, j}$ in.
\begin{remark}
\label{remark: expection}
 The ILVR~\cite{choi2021ilvr} method employs the low-pass filter to transfer the reference image information in Eqn.~\ref{equ: ILVR low pass}, and the low-pass filter calculates the block means. We have the following relationship between our mean restriction and the low-pass filter:
 \begin{equation}
  \mathbb{E}[\mathbf{\Phi}(\mathbf{y}_t)] = \otimes_{c,i,j}\sqrt{\bar{\alpha}_t}\operatorname{\mu}[\mathbf{y}_0^{c, i, j}]
 \end{equation}
\end{remark}

% We will introduce the Block Mean method of submanifold. We will give the formulation of design and fast way to calculate it. And show the improvement compared with ILVR.

% We will introduce our BAdaIN method of submanifold. We will give the formulation of design and fast way to calculate it. And show the improvement compared above method in theory, we will also show this in the experiments.

\textbf{Energy Function.}
% We will show our neural statistical design of the energy function, shallow VGG + BlockAdaIN + $L_2$ norm. Give the formulation and show it is a more stable way.
We can also use the BAdaIN module for constructing weak energy functions. Firstly, we use the first several layers of VGG19~\cite{simonyan2014very} net to extract neural features of $\mathbf{x}_t$ and $\mathbf{y}_t$ to get $\widehat{\mathbf{x}}_t$ and $\widehat{\mathbf{y}_t}$. Then we use the $L_2$ distance of $\operatorname{BAdaIN}(\widehat{\mathbf{x}}_t, \widehat{\mathbf{y}_t})$ and $\hat{\mathbf{x}_t}$ as the energy function for faithfulness. To verify SDDM's advantage, we only use this weak energy function.

%Xing：目前实验中的问题：缺少实验介绍和结果分析，不能只是罗列结果；另外还有很多语法错误和不规范排版
\section{Experiments}
\label{experiments}
\textbf{Datasets.} We evaluate our SDDM on the following datasets. All the images are resized to $256\times 256$ pixels.
\vspace{-4mm}
\begin{itemize}
\setlength{\itemsep}{0pt}
\setlength{\parsep}{0pt}
\setlength{\parskip}{3pt}
    \item AFHQ~\cite{Choi_2020_CVPR} contains high-resolution animal faces of three domains: cat, dog, and wild. Each domain has 500 testing images. we conduct Cat $\rightarrow$ Dog and Wild $\rightarrow$ Dog on this dataset, following the experiments of CUT~\cite{park2020contrastive} and EGSDE~\cite{zhao2022egsde}.
    \item CelebA-HQ~\cite{karras2017progressive} consists of high-quality human face images of two categories, male and female. Each category contains 1000 testing images. We conduct Male $\rightarrow$ Female on this dataset, following the experiments of EGSDE~\cite{zhao2022egsde}.
    \vspace{-3mm}
\end{itemize}

\textbf{Evaluation Metrics.} We evaluate our translated images from two aspects. One is to assess the distance between the translated and the source images, and we report the SSIM between them. The other is to evaluate the distance of generated images and target domain images, and we calculate the widely-used Frechet Inception Score (FID)~\cite{heusel2017gans} between the generated images and the target domain images. Details about the FID settings are in Appendix~\ref{sec: fid cal}.

\subsection{Comparison with the State-of-the-arts}
We compare our experiments with other GANs-based and SBDM-based methods, as shown in Table \ref{tab:quantitative comparison}.
%---------------------------Table-------------------------------------------------------------
\begin{table}[h]
\caption{Quantitative comparisons. The results marked with * come from~\cite{zhao2022egsde} Our method and ILVR have 100 diffusion steps. SDEdit and EGSDE* have 1000 diffusion steps. For a fair comparison with our SDDM, we also report the results of EGSDE** with 200 diffusion steps.  All SBDM-based methods are repeated 5 times to reduce the randomness. Details about SDDM and SDDM$^{\dagger}$ are shown in Appendix~\ref{section: details about settings}.} 
\label{tab:quantitative comparison}
\setlength\tabcolsep{7pt}
\vskip 0.05in
\begin{center}
\begin{small}
\begin{sc}
 \begin{tabular}{ccc}
  \toprule
  Model & FID$\downarrow$ & SSIM$\uparrow$\\
  \midrule
   &Cat $\rightarrow$ Dog & \\
  \midrule
  CycleGAN* &85.9 & -\\
  MUNIT* & 104.4 & -\\
  DRIT* & 123.4 & -\\
  Distance* & 155.3 & -\\
  SelfDistance* &144.4  & -\\
  GCGAN* & 96.6 & -\\
  LSeSim* &72.8  & -\\
  ITTR (CUT)* & 68.6 & -\\
   StarGAN v2* &\textbf{54.88 $\pm$ 1.01} & 0.27 ± 0.003 \\
  CUT* & 76.21 & \textbf{0.601} \\
  \midrule
  SDEdit* &74.17 ± 1.01 & \textbf{0.423 ± 0.001}\\
  ILVR* &74.37 ± 1.55 &0.363 ± 0.001 \\
  EGSDE* & 65.82 ± 0.77 & 0.415 ± 0.001\\
  EGSDE** & 70.16± 1.03& 0.411 ± 0.001\\
  SDDM(Ours) &\textbf{62.29 ± 0.63} & \textbf{0.422± 0.001} \\
  SDDM$^{\dagger}$ (Ours) &\textbf{49.43 ± 0.23} & 0.361± 0.001 \\
  \midrule
  % &Wild $\rightarrow$ Dog & \\
  % \midrule
  % CUT*&92.94 & 0.592\\
  % \midrule
  % SDEdit* & 68.51 ± 0.65 & 0.343 ± 0.001\\
  % ILVR* &75.33 ± 1.22 & 0.287 ± 0.001\\
  % EGSDE* &59.75 ± 0.62 & 0.343 ± 0.001\\
  % EGSDE** &&\\
  % SDDM(Ours) & & \\
  % \midrule
  &Wild $\rightarrow$ Dog & \\
  \midrule
  SDEdit* & 68.51 ± 0.65 & \textbf{0.343 ± 0.001}\\
  ILVR* & 75.33 ± 1.22 & 0.287 ± 0.001 \\
  EGSDE* & \textbf{59.75 ± 0.62} & \textbf{0.343 ± 0.001}\\
  SDDM(Ours) & \textbf{57.38 ± 0.53}& 0.328 ± 0.001\\
  
  \midrule
  % &Wild $\rightarrow$ Dog & \\
  % \midrule
  % CUT*&92.94 & 0.592\\
  % \midrule
  % SDEdit* & 68.51 ± 0.65 & 0.343 ± 0.001\\
  % ILVR* &75.33 ± 1.22 & 0.287 ± 0.001\\
  % EGSDE* &59.75 ± 0.62 & 0.343 ± 0.001\\
  % EGSDE** &&\\
  % SDDM(Ours) & & \\
  % \midrule
  &Male $\rightarrow$ Female & \\
  \midrule
  SDEdit* & 49.43 ± 0.47 &\textbf{0.572 ± 0.000 }\\
  ILVR* &46.12 ± 0.33 & 0.510 ± 0.001 \\
  EGSDE* &\textbf{41.93 ± 0.11} & \textbf{0.574 ± 0.000}\\
  EGSDE** & 45.12± 0.24& 0.512 ± 0.001 \\
  SDDM(Ours) & \textbf{44.37± 0.23}& 0.526 ± 0.001\\
  \bottomrule
 \end{tabular}
\end{sc}
\end{small}
\end{center}
\vskip -0.1in
\end{table}
%----------------------------------------------------------------------------------------------

Compared with other SBDM-based methods, our SDDM improves on both metrics FID and SSIM, which indicates the effectiveness of the two-stage generation process of our SDDM via the decomposition of score function and energy guidance with manifolds. Especially compared with EGSDE*, which has strong pre-trained energy functions, in the Cat $\rightarrow$ Dog task, our SDDM improves the FID score by 3.53 and SSIM score by 0.007 with much lower time steps, $1000\rightarrow 100$. For the comparison with EGSDE** having 200 diffusion steps, SDDM improves the FID score by 7.87 and the SSIM score by 0.011 in the Cat $\rightarrow$ Dog task and improves the FID score by 0.75 and the SSIM score by 0.014 in the Male $\rightarrow$ Female task, which suggests the advantage of our SDDM in fewer diffusion steps. The visual comparison is in Appendix~\ref{section: visual comparisons}.

\subsection{Ablation Studies}

\begin{figure}[htbp] %H为当前位置，!htb为忽略美学标准，htbp为浮动图形
\centering %图片居中
\includegraphics[width=.9\columnwidth]{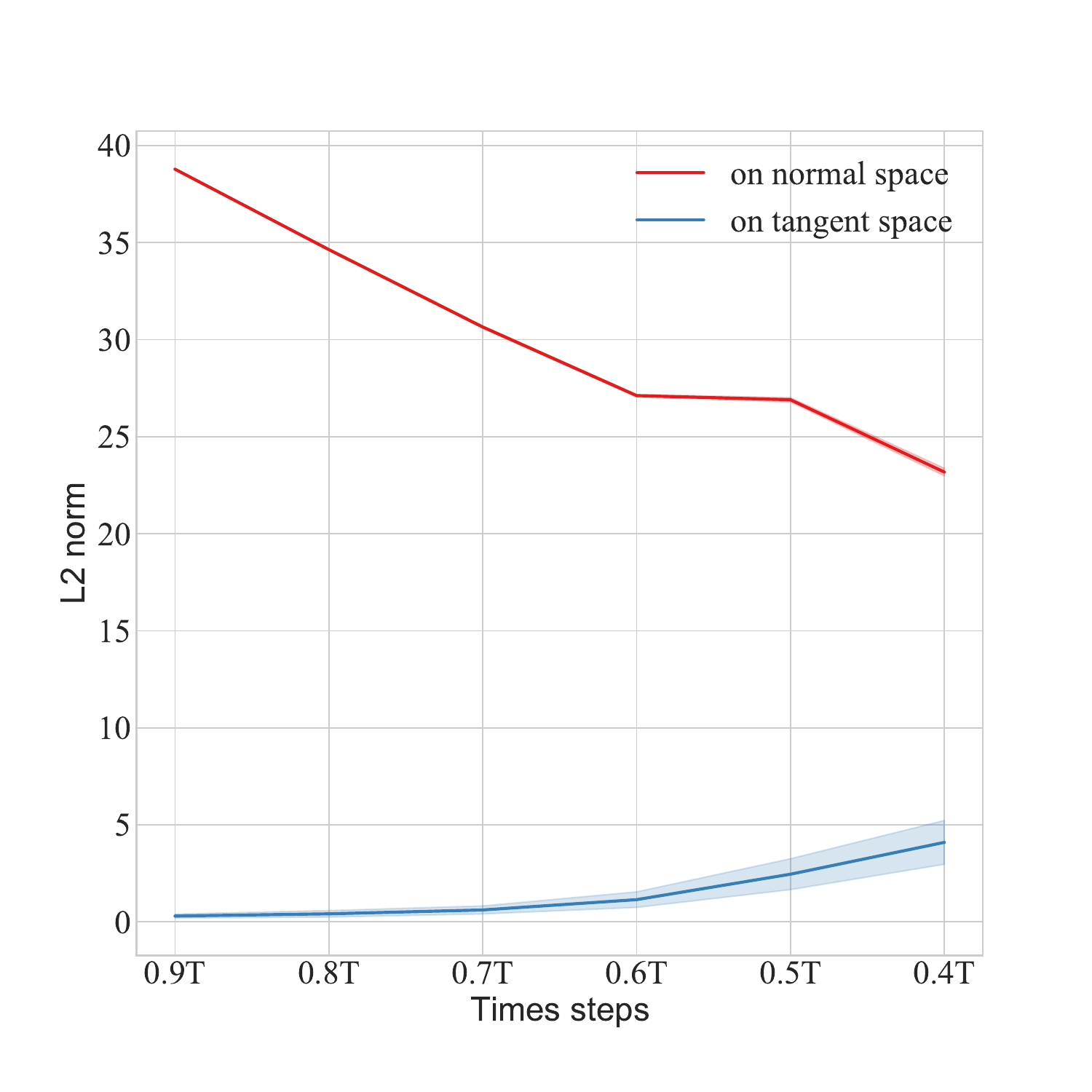} %插入图片，[]中设置图片大小，{}中是图片文件名
\vspace{-0.8cm}
\caption{The mean and standard deviation of $L_2$ norms of $\mathbf{s}_r$ in $T_{\mathbf{x}_t} \mathcal{M}$ and other part in $N_{\mathbf{x}_t} \mathcal{M}$. We repeat 100 times of our SDDM with different reference images.} %最终文档中希望显示的图片标题
\label{Fig.percentage} %用于文内引用的标签
\end{figure}

\textbf{Observations on Score Components.}
% 画成图
While performing the Cat $\rightarrow$ Dog experiment, we report the $L_2$ norms of the deterministic guidance values on $T_{\mathbf{x}_t} \mathcal{M}$ and $N_{\mathbf{x}_t} \mathcal{M}$. As shown in Figure~\ref{Fig.percentage}, the component on the normal space has one in 128 dimensions but contains the most value of the deterministic guidance of diffusion models, while the component on the tangent space $\mathbf{s}_r(\cdot, \cdot)$ has 127 in 128 dimensions but contains a minimal value, which indicates we have relative large optimization space on the manifold which will not excessively interfere with the intermediate distributions.
  
% TODO IMAGE
% \begin{table}[htbp]
%  \centering
%  \resizebox{\columnwidth}{!}{
%  \begin{tabular}{ccccccc}
%   \toprule
%    Space & Dimensions & 0.9T & 0.8T& 0.7T & 0.6T & 0.5T\\
%    \midrule
%    $T_{\mathbf{x}_t} \mathcal{M}$ &127/128& 0.3777 & 0.3681 & 0.8434 & 1.6634 & 2.5837 \\
%    $N_{\mathbf{x}_t} \mathcal{M}$ &1/128& 77.5988 & 69.3881 & 60.7217 & 54.4680 & 49.8738 \\
%    \bottomrule
%   \end{tabular}}
%   \caption{$L_2$ norm of different parts, This will becomes a picture.} 
%   \label{tab:l2norm}
%   \end{table}

\textbf{Comparison of Different Manifolds.}
We compare SDDM with different manifold methods and report the results in Table~\ref{tab: compare manifolds}. Compared with the manifold restricted with a low-pass filter, the manifold restricted with our BAdaIN has better performance both on FID and SSIM, because our manifold separates the content refinement part and image denoising part better.

%---------------------------Table-------------------------------------------------------------
\begin{table}[htbp]
\caption{Comparisons of different manifolds.} 
\label{tab: compare manifolds}
\setlength\tabcolsep{14pt}
\vskip 0.05in
\begin{center}
\begin{small}
\begin{sc}
\begin{tabular}{lcc}
\toprule
Model & FID$\downarrow$ & SSIM$\uparrow$\\
\midrule
% SDDM(low-pass filter) &74.37&56.95 &  17.77 & 0.363\\
% SDDM(low-pass filter) &67.79& \textbf{51.49} & \textbf{18.81} & \textbf{0.420}\\
SDDM(low-pass filter) & 67.56  & 0.411\\
SDDM(BAdaIN) &\textbf{62.29} & \textbf{0.422} \\
\bottomrule
\end{tabular}
\end{sc}
\end{small}
\end{center}
\vskip -0.1in
\end{table}
%----------------------------------------------------------------------------------------------

\textbf{Comparison of Different Coefficients.}
We have two coefficients at each iteration step, the coefficient of the step size $\lambda$ of optimal multi-objection direction and the coefficient of the energy guidance $\lambda_1$. As in Table~\ref{tab: compare coefficients}, the larger $\lambda$ is, the better FID will be because, at each optimization on the manifold, it reaches a position with higher probability $p_{\mathcal{M}}(\mathbf{x})$, but when $\lambda$ is too large, the FID score will be worse again. The $\lambda_1$ has a negative connection with the impact of energy guidance, which indicates that smaller $\lambda_1$ makes the energy guidance stronger and thus has a better SSIM score.

%---------------------------Table-------------------------------------------------------------
\begin{table}[htbp]
\caption{Comparisons of different coefficients.}
\label{tab: compare coefficients}
\setlength\tabcolsep{20pt}
\vskip 0.05in
\begin{center}
\begin{small}
\begin{sc}
\begin{tabular}{lcc}
    \toprule
    Coefficients & FID$\downarrow$ & SSIM$\uparrow$\\
    \midrule
% SDDM(low-pass filter) &74.37&56.95 &  17.77 & 0.363\\
% SDDM(low-pass filter) &67.79& \textbf{51.49} & \textbf{18.81} & \textbf{0.420}\\
$\lambda = 2, \lambda_1 = 10$ & 65.09 & \textbf{0.429}\\
$\lambda = 2, \lambda_1 = 40$ & \textbf{62.02} & 0.420\\
$\lambda = 1, \lambda_1 = 25$ & 66.04 & 0.428\\
$\lambda = 3, \lambda_1 = 25$ & 62.32 & 0.415\\
$\lambda = 2, \lambda_1 = 25$ &62.29 & 0.422 \\
\bottomrule
\end{tabular}
\end{sc}
\end{small}
\end{center}
\vskip -0.1in
\end{table}
%----------------------------------------------------------------------------------------------

\textbf{Comparison w./w.o. Multi-Objective Optimization.}
We compare SDDM with SDDM without the MOO method and report the FID, SSIM, and probability of negative impact (PNI), which indicates the situation that the total guidance including score and energy decreases the $p(\mathbf{x})$ in Table~\ref{tab:copmare moo}. The proposed SDDM method avoids such situations and reaches better performance.

%---------------------------Table-------------------------------------------------------------
\begin{table}[htbp]
\vspace{-3mm}
\caption{Comparisons of different manifold optimization policies.} 
\label{tab:copmare moo}
\setlength\tabcolsep{11pt}
\vspace{-3mm}
\begin{center}
\begin{small}
\begin{sc}
\begin{tabular}{lccc}
  \toprule
  Model & FID$\downarrow$& SSIM$\uparrow$ & PNI$\downarrow$\\
  \midrule
% SDDM(low-pass filter) &74.37&56.95 &  17.77 & 0.363\\
SDDM(w/o MOO)  & 64.93 & 0.421 & 0.024\\
SDDM &\textbf{62.29} & \textbf{0.422} & \textbf{0}\\
\bottomrule
\end{tabular}
\end{sc}
\end{small}
\end{center}
\vspace{-1mm}
\end{table}
%----------------------------------------------------------------------------------------------
\vspace{-3mm}

\textbf{$\epsilon$ Policy in The Optimization on Manifolds.}
We mainly compare three different $\epsilon$ policies:
\vspace{-2mm}
\begin{itemize}
\setlength{\itemsep}{0pt}
\setlength{\parsep}{0pt}
\setlength{\parskip}{3pt}
    \item Policy 1: The $\epsilon$ is very small such that in Algorithm~\ref{alg:opt on manifold}, we iterate only once. With Remark~\ref{remark: simulate}, this method will introduce no additional calculations of scores.
    \item Policy 2: $\epsilon_t = \|\mathbf{s}_r(\mathbf{x}, t)\|$, which normally iterates twice in Algorithm~\ref{alg:opt on manifold}. In practice, we iterate twice.
    \item Policy 3: At the time step $T_0$ in Algorithm~\ref{alg:generation_with_sddm}, we set $\epsilon_t$ larger to iterate another 4 times. and other time steps are as same as Policy 1.
\vspace{-2mm}
\end{itemize}
We report the FID and SSIM of different policies in Table~\ref{tab:copmare policy}. Policy 3 has the best performance, which reveals that iteration a little more at $T_0$ time step can balance different metrics better without introducing too much cost.
%---------------------------Table-------------------------------------------------------------
\begin{table}[htbp]
\vspace{-2mm}
\caption{Comparisons of the $\epsilon$ Policies.} 
\label{tab:copmare policy}
\setlength\tabcolsep{23pt}
\vskip 0.05in
\begin{center}
\begin{small}
\begin{sc}
\begin{tabular}{lcc}
  \toprule
  Policy & FID$\downarrow$& SSIM$\uparrow$\\
  \midrule
% SDDM(low-pass filter) &74.37&56.95 &  17.77 & 0.363\\
Policy 1  & \textbf{61.33} &0.413 \\
Policy 2  & 64.05 & 0.418 \\
Policy 3 & 62.29 & \textbf{0.422} \\
\bottomrule
\end{tabular}
\end{sc}
\end{small}
\end{center}
\vskip -0.1in
\end{table}
%----------------------------------------------------------------------------------------------

\begin{figure}[htbp] %H为当前位置，!htb为忽略美学标准，htbp为浮动图形
  \centering %图片居中
  \includegraphics[width=.9\columnwidth]{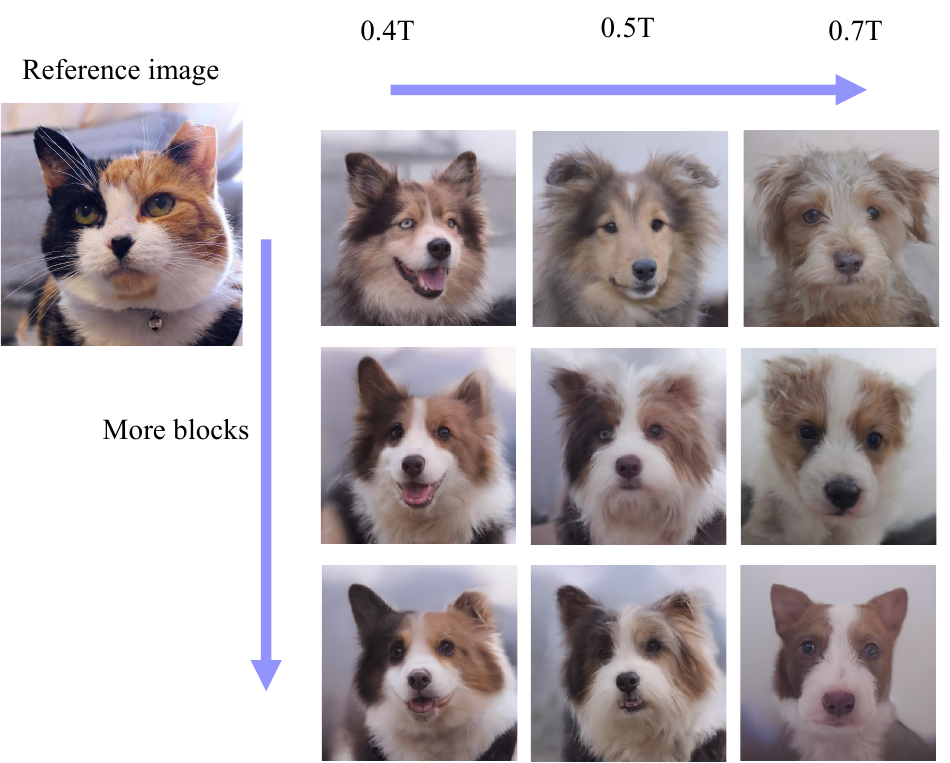} %插入图片，[]中设置图片大小，{}中是图片文件名
  \caption{The comparison of different numbers of blocks and different middle time $T_0$s.} %最终文档中希望显示的图片标题
  \label{Fig.block and time} %用于文内引用的标签
  \end{figure}

\textbf{The Choice of Middle-Time $T_0$ and Block Number.}
As shown in Figure~\ref{Fig.block and time}, when we chunk more blocks or set the $T_0$ smaller, the generated image is more faithful to the reference image. But too many blocks will also introduce some bad details, like the mouth in the left bottom image.

%---------------------------Table-------------------------------------------------------------
\begin{table}[htbp]
\vspace{-2mm}
\caption{Comparisons of different block numbers.} 
\label{tab:copmare blocknumbers}
\setlength\tabcolsep{19pt}
\vskip 0.05in
\begin{center}
\begin{small}
\begin{sc}
\begin{tabular}{lcc}
  \toprule
  Block number & FID$\downarrow$& SSIM$\uparrow$\\
  \midrule
% SDDM(low-pass filter) &74.37&56.95 &  17.77 & 0.363\\
$8 \times 8$  & \textbf{54.56} & 0.359\\
$16 \times 16 $  & 62.29 & 0.422 \\
$32\times 32$ & 68.03 & \textbf{0.426} \\
\bottomrule
\end{tabular}
\end{sc}
\end{small}
\end{center}
\vskip -0.1in
\end{table}
%----------------------------------------------------------------------------------------------

\section{Related Work}
\textbf{GAN-based Unpaired Image-to-Image Translation}.
There are mainly two categories of GANs-based methods in the unpaired I2I task. One is two-side way, while the other is one-side mapping. In the first category, the key idea is that the translated image could be translated back with another inverse mapping. CycleGAN~\cite{zhu2017unpaired}, DualGAN~\cite{yi2017dualgan}, DiscoGAN~\cite{kim2017learning}, SCAN~\cite{van2020scan} and U-GAT-IT~\cite{kim2019u} are in this class. But translations usually lose information. Several new studies have started to map two domains to the same metric space and use the distance of this space as supervision.DistanceGAN~\cite{benaim2017one}, GCGAN~\cite{fu2019geometry}, CUT~\cite{park2020contrastive} and LSeSim~\cite{zheng2021spatially} are in this categoriy.

It is also noteworthy that other techniques have been proposed to tackle the problem of unpaired image-to-image translation. For instance, some studies~\cite{xie2021learning, xie2018cooperative} leverage cooperative learning, whereas others~\cite{zhao2021learning} adopt an energy-based framework or a short-run MCMC like Langevin dynamics~\cite{xie2016theory}.
% \subsection{Diffusion Models}
% Several discrete diffusion models are proposed first, including diffusion probabilistic models~\cite{sohl2015deep}, noise-conditioned score network, denoising diffusion probabilistic models~\cite{Ho2020DDPM}. \cite{song2020score} find the forward of all these diffusion processes can be described with stochastic diffusion equations, and  more surprisingly, \cite{anderson1982reverse} have proved the reverse process can be simulated out with the score function. Then they use the pointview of the score function to unify all the diffusion models.

\textbf{SBDMs-based Conditional Methods}.
There are mainly two classes of conditional generation with SBDMs. The first one is to empower SBDMs with the conditional generation ability during training with the classifier-free guidance trick~\cite{ho2022classifier}, which learns the score functions and conditional score functions via a single neural network. The other method is to train another classifier to lead the learned score functions for a conditional generation. EGSDE~\cite{zhao2022egsde} generalizes the classifiers to any energy-based functions. These methods cannot describe the intermediate distributions clearly, which is a hard problem because the distributions of adjacent time steps are deeply coupled. However, when the conditions can give constraints to separate the adjacent distributions well, we can get better results, and this observation inspires our model.

\section{Conclusions}
\label{conclusion}
In this work, we have presented a new score-decomposed diffusion model, SDDM, which leverages manifold analyses to decompose the score function and explicitly optimize the tangled distributions during image generation. SDDM derives manifolds to separate the distributions of adjacent time steps and decompose the score function or energy guidance into an image “denoising” part and a content “refinement” part. With the new multi-objective optimization algorithm and block adaptive instance normalization module, our realized SDDM method demonstrates promising results in unpaired image-to-image translation on two benchmarks. In future work, we plan to improve and apply the proposed SDDM model in more image translation tasks.

One limitation of our approach involves additional computations, although these computations are negligible compared to the inferences of neural networks. Additionally, we should prevent any misuse of generative algorithms for malicious purposes.

% % Acknowledgements should only appear in the accepted version.
\section*{Acknowledgements}
This work is supported by the National Key R\&D Program of China under Grant No. 2021QY1500, and the State Key Program of the National Natural Science Foundation of China (NSFC) (No.61831022). It is also partly supported by the NSFC under Grant No. 62076238 and 62222606.
% \textbf{Do not} include acknowledgments in the initial version of
% the paper submitted for blind review.

% If a paper is accepted, the final camera-ready version can (and
% probably should) include acknowledgments. In this case, please
% place such acknowledgments in an unnumbered section at the
% end of the paper. Typically, this will include thanks to reviewers
% who gave useful comments, to colleagues who contributed to the ideas,
% and to funding agencies and corporate sponsors that provided financial
% support.

% In the unusual situation where you want a paper to appear in the
% references without citing it in the main text, use \nocite
% \nocite{langley00}

% xing：参考文献的格式急需要统一，包括大小写、是否缩写、包含条目是否一致等。
\bibliography{SDDM}
\bibliographystyle{icml2023}

%%%%%%%%%%%%%%%%%%%%%%%%%%%%%%%%%%%%%%%%%%%%%%%%%%%%%%%%%%%%%%%%%%%%%%%%%%%%%%%
%%%%%%%%%%%%%%%%%%%%%%%%%%%%%%%%%%%%%%%%%%%%%%%%%%%%%%%%%%%%%%%%%%%%%%%%%%%%%%%
% APPENDIX
%%%%%%%%%%%%%%%%%%%%%%%%%%%%%%%%%%%%%%%%%%%%%%%%%%%%%%%%%%%%%%%%%%%%%%%%%%%%%%%
%%%%%%%%%%%%%%%%%%%%%%%%%%%%%%%%%%%%%%%%%%%%%%%%%%%%%%%%%%%%%%%%%%%%%%%%%%%%%%%
\newpage
\appendix
\onecolumn
\section{Basic Knowledge about Manifolds}
\label{app: BK about manifolds}
These definitions come from~\cite{lee2010introduction, tu2011manifolds, lee2012smooth}.

\begin{definition}\textbf{Topological space.}

  A topological space $\mathcal{M}$ is locally Euclidean of dimension $n$ if every point $p$ in $\mathcal{M}$ has a neighborhood $U$ such that there is a homeomorphism $\phi$ from $U$ onto an open subset of $\mathbb{R}^n$. We call the pair $(U, \phi: U  \rightarrow \mathbb{R}^n)$ a chart, $U $ a coordinate neighborhood or an open coordinate set, and $\phi$ a coordinate map or a coordinate system on $U$. We say that a chart $(U, \phi)$ is centered at $p \in U$ if $\phi(p) = 0$. A chart $(U, \phi)$ about $p$ simply means that $(U, \phi)$ is a chart and $p \in U$ .
\end{definition}

\begin{definition}\textbf{Locally Euclidean property.}

  The locally Euclidean property means that for each $p \in \mathcal{M} $, we can find the following:
  \begin{itemize}
    \item an open set $U \subset \mathcal{M}$ containing $p$;
    \item an open set $ \widetilde{U} \subset \mathbb{R}^n$;and
    \item a homeomorphism $\phi : U \rightarrow \widetilde{U}$ (\textit{i.e.}, a continuous bijective map with continuous inverse).
  \end{itemize}
\end{definition}

\begin{definition}\textbf{Topological manifold.}

Suppose $\mathcal{M}$ is a topological space. We say $\mathcal{M}$ is a topological manifold of dimension $n$ or a topological $n$-manifold if it has the following properties:
\begin{itemize}
  \item M is a Hausdorff space: For every pair of points $p, q \in \mathcal{M}$, there are disjoint open subsets $U, V \subset \mathcal{M}$ such that $p \in U $and $q \in V$.
  \item $\mathcal{M}$ is second countable: There exists a countable basis for the topology of $\mathcal{M}$.
  \item $\mathcal{M}$ is locally Euclidean of dimension $n$: Every point has a neighborhood that is homeomorphic to an open subset of $\mathbb{R}^n$.
\end{itemize}
\end{definition}

\begin{definition}\textbf{Tangent vector.}

  A tangent vector at a point $p$ in a manifold $\mathcal{M}$ is a derivation at $p$.
\end{definition}

\begin{definition}\textbf{Tangent space.}

  As for $\mathbb{R}^n$, the tangent vectors at $p$ form a vector space $T_p(\mathcal{M})$, called the tangent space of $\mathcal{M}$ at $p$. We also write $T_p\mathcal{M}$ instead of $T_p(\mathcal{M})$.
\end{definition}

\begin{definition}\textbf{Normal space.}
  
  the normal space to  $\mathcal{M}$ at $p$ to be the subspace $N_p\mathcal{M} \subset  \mathbb{R}^m$ consisting of all vectors that are orthogonal to $T_p\mathcal{M}$ with respect to the Euclidean dot product. The normal bundle of M is the subset  $N \mathcal{M} \subset  \mathbb{R}^m \times \mathbb{R}^m$defined by
  \begin{equation}
    N \mathcal{M}=\coprod_{p \in M} N_p \mathcal{M}=\left\{(p, v) \in \mathbb{R}^m \times \mathbb{R}^m: p \in M \text { and } v \in N_p \mathcal{M}\right\}
    \end{equation}
\end{definition}

\section{Proofs}

\label{app: proofs}
\subsection{Proof of Lemma~\ref{lemma: decouple of score}}
\textit{Proof.} 
Consider the local coordinate system at $\mathbf{x}$. Suppose $\{x_i\}_{i = 1, 2, ..., d}$ are the orthonormal basis of $\mathbb{R}^d$ and $\{x_i\}_{i = 1, 2, ..., m}$ ($m < d$) are in the tangent space $T_{\mathbf{x}}\mathcal{M}$ and the rest of them are in the normal space $N_{\mathbf{x}}\mathcal{M}$. Then:
\begin{equation}
  \begin{aligned}
    \mathbf{s}_r(\mathbf{x})=&\nabla_{\mathbf{x}}\log p_{\mathcal{M}}(\mathbf{x})
    =\sum_{i = 1}^{d}\frac{\partial}{\partial x_i}\log p_{\mathcal{M}}(\mathbf{x})\\
    =&\sum_{i = 1}^{m}\frac{\partial}{\partial x_i}\log p_{\mathcal{M}}(\mathbf{x})
    =\sum_{i = 1}^{m}\frac{\partial}{\partial x_i}\log C p(\mathbf{x})\\
    =&\sum_{i = 1}^{m}\frac{\partial}{\partial x_i}\log p(\mathbf{x})
    =\mathbf{s}(\mathbf{x})|_{T_{\mathbf{x}}\mathcal{M}}.
  \end{aligned}
\end{equation}

Therefore, we have:
\begin{equation}
  \begin{aligned}
    \mathbf{s}(\mathbf{x})=&\mathbf{s}(\mathbf{x})|_{T_{\mathbf{x}}\mathcal{M}} + \mathbf{s}(\mathbf{x})|_{N_{\mathbf{x}}\mathcal{M}}\\
    =&\mathbf{s}_r(\mathbf{x}) + \mathbf{s}_d(\mathbf{x})
  \end{aligned}
\end{equation}
\hfill \ensuremath{\Box}

In the following sections, consider the distributions of perturbed reference image $\mathbf{y_t} = \hat{\alpha}_t\mathbf{y}_0 + \hat{\beta}_t \mathbf{z}_t$, where $\mathbf{z}_t \sim \mathcal{N}(\mathbf{0}, \mathbf{I})$, and the reference image $\mathbf{y}_0$ is fixed.

\subsection{Proof of Proposition~\ref{prop: compact_manifolds} and Lemma~\ref{lemma:detailed_compact_manifold}}

Before proving the Proposition~\ref{prop: compact_manifolds} and Lemma~\ref{lemma:detailed_compact_manifold}, we will prove another two relevant lemmas.
\begin{lemma}
  \label{lemma: meanrestriction}
  $\mathbf{y_t}$ is clustered on the $(d-1)-dim$ manifolds $\{\mathcal{M}_t\}$ restricted with the first-order moment constraints, 
  \begin{equation}
    \label{equ: mean restrictions}
    \operatorname{\mu}[\mathbf{x}_t] = \hat{\alpha}_t \operatorname{\mu}[\mathbf{y}_0]
  \end{equation}
  under the $d_2$ distance of $\mathbb{R}^d$. 
  
  Strictly speaking,  in the original Cartesian coordinate system $\mathbb{R}^{d}$.

  $\forall \epsilon, \xi > 0, \exists D >0, \forall d > D$ we have:
  
  $$\mathcal{P}\left(d_2\left(\mathbf{y}_t, \mathcal{M}_t\right) <\epsilon \sqrt{d}\right) > 1 - \xi $$
\end{lemma}

\textit{Proof.} 
The manifold provided with restriction of Eqn.~\eqref{equ: mean restrictions} is a hyperplane in $\mathbb{R}^d$, and the normal vector $\mathbf{n} =  \frac{1}{\sqrt{d}}(1,1,...,1)$. Then the $L_2$ distance from $\mathbf{y}_t$ to the manifold $\mathcal{M}_t$ is $| \hat{\beta}_t \mathbf{z}_t \cdot \mathbf{n} |$, where $\hat{\beta}_t \mathbf{z}_t \cdot \mathbf{n} \sim \mathcal{N}(0, \hat{\beta}_t^2)$. Therefore, 
\begin{equation}
  \begin{aligned}
    d_2\left(\mathbf{y}_t, \mathcal{M}_t\right) = \sqrt{d} | \frac{1}{\sqrt{d}} \hat{\beta}_t \mathbf{z}_t \cdot \mathbf{n} |
  \end{aligned}
\end{equation}
Thus, as $d \rightarrow +\infty$, the variance of $\frac{1}{\sqrt{d}}  \hat{\beta}_t \mathbf{z}_t \cdot \mathbf{n} \rightarrow 0$. 

Then strictly speaking,   $\forall \epsilon, \xi > 0, \exists D >0, \forall d > D$ we have:
  
$$\mathcal{P}\left(d_2\left(\mathbf{y}_t, \mathcal{M}_t\right) <\epsilon \sqrt{d}\right) > 1 - \xi $$
\hfill \ensuremath{\Box}

\begin{lemma}
  \label{lemma:variancerestriction}
  Suppose $\mathbf{y}_t$ shares the same bound $A$, which means $\|\mathbf{y}_t\|_{\infty} < A$.
$\mathbf{y_t}$ is clustered on the $(d-1)-dim$  manifolds $\{\mathcal{M}_t\}$ restricted with the second-order moment constraints, 
  \begin{equation}
    \label{equ: var restrictions}
    \operatorname{\mu}\left[\mathbf{x}_t - \hat{\alpha}_t \operatorname{\mu}[\mathbf{y}_0]\right]^2 = \hat{\alpha}^2_t\operatorname{Var}[\mathbf{y}_0] + \hat{\beta}^2_t
  \end{equation}
  under the metric of $d_2$ distance.

  Strictly speaking,

  $\forall \epsilon, \xi > 0, \exists D >0, \forall d > D$ we have:
  
  $$\mathcal{P}\left(d_2\left(\mathbf{y}_t, \mathcal{M}_t\right) <\epsilon \sqrt{d}\right) > 1 - \xi $$
\end{lemma}

\textit{Proof.} 
The manifold provided with restriction of Eqn.~\eqref{equ: var restrictions} is a hypersphere on the $\mathbb{R}^d$. The center of the hypersphere is $\hat{\alpha}_t \operatorname{\mu}[\mathbf{y}_0] (1,1,...,1)$ and the radius is $\sqrt{d} \sqrt{\hat{\alpha}^2_t\operatorname{Var}[\mathbf{y}_0] + \hat{\beta}^2_t}$. The square of the $L_2$ distance of $\mathbf{y}_t$ to the center is:
\begin{equation}
  \begin{aligned}
    \left[\hat{\alpha}_t\mathbf{y}_0 + \hat{\beta}_t \mathbf{z}_t - \hat{\alpha}_t \operatorname{\mu}[\mathbf{y}_0]\right]^2 &= \sum_{i = 1}^d  \left[(\hat{\alpha}_t\mathbf{y}_0^i - \hat{\alpha}_t \operatorname{\mu}[\mathbf{y}_0]) + \hat{\beta}_t \mathbf{z}_t^i \right]^2 \\ 
    &= \sum_{i = 1}^d \left[(\hat{\alpha}_t\mathbf{y}_0^i - \hat{\alpha}_t \operatorname{\mu}[\mathbf{y}_0])^2 + \hat{\beta}_t^2 (\mathbf{z}_t^i)^2 + 2(\hat{\alpha}_t\mathbf{y}_0^i - \hat{\alpha}_t \operatorname{\mu}[\mathbf{y}_0])\hat{\beta}_t \mathbf{z}_t^i \right] \\
    & = d \hat{\alpha}^2_t\operatorname{Var}[\mathbf{y}_0] + \hat{\beta}_t^2 \mathbf{z}_t^2 + 2 \hat{\alpha}_t \hat{\beta}_t (\mathbf{y_0} - \operatorname{\mu}[\mathbf{y}_0])\cdot \mathbf{z}_t, 
  \end{aligned}
\end{equation}

Therefore,
\begin{equation}
  \begin{aligned}
    d_2\left(\mathbf{y}_t, \mathcal{M}_t\right) =& \left|\sqrt{\left[\hat{\alpha}_t\mathbf{y}_0 + \hat{\beta}_t \mathbf{z}_t - \hat{\alpha}_t \operatorname{\mu}[\mathbf{y}_0]\right]^2} -  \sqrt{d} \sqrt{\hat{\alpha}^2_t\operatorname{Var}[\mathbf{y}_0] + \hat{\beta}^2_t} \right| \\
    =& \frac{\left|\left[\hat{\alpha}_t\mathbf{y}_0 + \hat{\beta}_t \mathbf{z}_t - \hat{\alpha}_t \operatorname{\mu}[\mathbf{y}_0]\right]^2 - d \hat{\alpha}^2_t\operatorname{Var}[\mathbf{y}_0] - d\hat{\beta}^2_t\right| }{ \sqrt{\left[\hat{\alpha}_t\mathbf{y}_0 + \hat{\beta}_t \mathbf{z}_t - \hat{\alpha}_t \operatorname{\mu}[\mathbf{y}_0]\right]^2} +  \sqrt{d} \sqrt{\hat{\alpha}^2_t\operatorname{Var}[\mathbf{y}_0] + \hat{\beta}^2_t}}\\
    \le& \frac{1}{\sqrt{d} \hat{\beta}_t}\left|\left[\hat{\alpha}_t\mathbf{y}_0 + \hat{\beta}_t \mathbf{z}_t - \hat{\alpha}_t \operatorname{\mu}[\mathbf{y}_0]\right]^2 - d \hat{\alpha}^2_t\operatorname{Var}[\mathbf{y}_0] - d \hat{\beta}^2_t\right| \\
    =& \frac{1}{\sqrt{d} \hat{\beta}_t} \left| d \hat{\alpha}^2_t\operatorname{Var}[\mathbf{y}_0] + \hat{\beta}_t^2 \mathbf{z}_t^2 + 2 \hat{\alpha}_t \hat{\beta}_t (\mathbf{y_0} - \operatorname{\mu}[\mathbf{y}_0])\cdot \mathbf{z}_t - d \hat{\alpha}^2_t \operatorname{Var}[\mathbf{y}_0] - d \hat{\beta}^2_t \right| \\
    \le& \frac{1}{\sqrt{d}}\left(\hat{\beta}_t^2\left| \mathbf{z}_t^2 - d\right| + \left|2 \hat{\alpha}_t \hat{\beta}_t (\mathbf{y_0} - \operatorname{\mu}[\mathbf{y}_0])\cdot \mathbf{z}_t\right| \right)\\
    =& \sqrt{d}\left(\hat{\beta}_t\frac{\left| \mathbf{z}_t^2 - d\right|}{d} + 2 \hat{\alpha}_t \frac{\left|(\mathbf{y_0} - \operatorname{\mu}[\mathbf{y}_0])\cdot \mathbf{z}_t\right|}{d} \right), \\
  \end{aligned}
\end{equation}
where the $\mathbf{z}_t^2$ is a stand chi-square distribution with $d$ degrees of freedom. We apply the standard Laurent-Massart bound~\cite{laurent2000adaptive} for it and get 

\begin{equation}
  \begin{aligned}
  \mathcal{P}[\mathbf{z}_t^2- d \geq 2 \sqrt{d t}+2 t] & \leq e^{-t} \\
  \mathcal{P}[\mathbf{z}_t^2-d \leq-2 \sqrt{d t}] & \leq e^{-t},
  \end{aligned}
  \end{equation}
which holds for any $t > 0$.
We let $ t = d \epsilon'$, where the $\epsilon'+ \sqrt{\epsilon'} = \frac{\epsilon}{4\hat{\beta}_t}$ for any given $\epsilon$, Then we have 
\begin{equation}
  \mathcal{P}\left[ -2d \sqrt{\epsilon'} \le \mathbf{z}_t^2 - d \le 2d (\epsilon'+ \sqrt{\epsilon'})\right] \ge 1 - 2e^{-d\epsilon'}.
\end{equation}
Therefore, 
\begin{equation}
  \mathcal{P}\left[\hat{\beta}_t\frac{\left| \mathbf{z}_t^2 - d\right|}{d} < \frac{\epsilon}{2}\right] > 1 - 2e^{-d\epsilon'}
\end{equation}
and $\exists D_1, \forall d > D_1,  4e^{-d\epsilon'} \le \xi $, thus 

\begin{equation}
  \mathcal{P}\left[\hat{\beta}_t\frac{\left| \mathbf{z}_t^2 - d\right|}{d} < \frac{\epsilon}{2}\right] > 1 - \frac{\xi}{2}
\end{equation}

Similar to Lemma~\ref{lemma: meanrestriction}, $2 \hat{\alpha}_t \frac{(\mathbf{y_0} - \operatorname{\mu}[\mathbf{y}_0])\cdot \mathbf{z}_t}{d}$ is a Gaussian distribution, and the mean is 0, variance is bounded with $\frac{(4 \hat{\alpha}_t A)^2}{d} $. As $d \rightarrow +\infty$, the variance $\rightarrow 0$, thus $\exists D_2, \forall d > D_2$, we have
\begin{equation}
  \mathcal{P}\left[2 \hat{\alpha}_t \frac{\left|(\mathbf{y_0} - \operatorname{\mu}[\mathbf{y}_0])\cdot \mathbf{z}_t\right|}{d} < \frac{\epsilon}{2}\right] > 1 - \frac{\xi}{2} 
\end{equation}

Finally,  $\forall \epsilon, \xi > 0, \exists D = Max\{D_1, D_2\}, \forall d > D$, 
\begin{equation}
  \begin{aligned}
    \mathcal{P}\left(d_2\left(\mathbf{y}_t, \mathcal{M}_t\right) <\epsilon \sqrt{d}\right) \ge& \mathcal{P}\left[\hat{\beta}_t\frac{\left| \mathbf{z}_t^2 - d\right|}{d} < \frac{\epsilon}{2},2 \hat{\alpha}_t \frac{\left|(\mathbf{y_0} - \operatorname{\mu}[\mathbf{y}_0])\cdot \mathbf{z}_t\right|}{d} < \frac{\epsilon}{2} \right] \\
    =& \mathcal{P}\left[\hat{\beta}_t\frac{\left| \mathbf{z}_t^2 - d\right|}{d} < \frac{\epsilon}{2} \right] + \mathcal{P}\left[2 \hat{\alpha}_t \frac{\left|(\mathbf{y_0} - \operatorname{\mu}[\mathbf{y}_0])\cdot \mathbf{z}_t\right|}{d} < \frac{\epsilon}{2} \right] \\ &- \mathcal{P}\left[\hat{\beta}_t\frac{\left| \mathbf{z}_t^2 - d\right|}{d} < \frac{\epsilon}{2} \text{ or } 2 \hat{\alpha}_t \frac{\left|(\mathbf{y_0} - \operatorname{\mu}[\mathbf{y}_0])\cdot \mathbf{z}_t\right|}{d} < \frac{\epsilon}{2} \right]\\
    \ge& 1- \frac{\xi}{2} +  1- \frac{\xi}{2} - 1 \\
    =& 1 - \xi
  \end{aligned}
\end{equation}
\hfill \ensuremath{\Box}

Then we will prove the Proposition~\ref{prop: compact_manifolds}.

\textit{Proof.}
Consider $\mathcal{M}_t$, which is restricted with:
\begin{equation}
  \begin{aligned}
   \operatorname{\mu}[\mathbf{x}_t] &= \hat{\alpha}_t \operatorname{\mu}[\mathbf{y}_0],\\
   \operatorname{Var}[\mathbf{x}_t] &= \hat{\alpha}^2_t\operatorname{Var}[\mathbf{y}_0] + \hat{\beta}^2_t.
  \end{aligned}
 \end{equation}
We can substitute the $\operatorname{\mu}[\mathbf{x}_t]$ with $\sqrt{\bar{\alpha}_t}\operatorname{\mu}[\mathbf{y}_0]$ in the calculation of the variance and get the following equivalent restrictions:

\begin{equation}
\label{eqn. for define separated manifolds}
  \begin{aligned}
   \operatorname{\mu}[\mathbf{x}_t] &= \hat{\alpha}_t \operatorname{\mu}[\mathbf{y}_0],\\
   \operatorname{\mu}\left[\mathbf{x}_t - \hat{\alpha}_t \operatorname{\mu}[\mathbf{y}_0]\right]^2&= \hat{\alpha}^2_t\operatorname{Var}[\mathbf{y}_0] + \hat{\beta}^2_t.
  \end{aligned}
 \end{equation}

These two constraints correspond to Lemma~\ref{lemma: meanrestriction} and Lemma~\ref{lemma:variancerestriction} respectively.
We denote the manifold restricted with one of these constraints as $\mathcal{M}_{tA}$ and $\mathcal{M}_{tB} $. $\mathcal{M}_{tA} \cap \mathcal{M}_{tB} = \mathcal{M}_t$. Suppose the angle of $\mathcal{M}_{tA}$ and $\mathcal{M}_{tB} $ at the intersection is $\theta$. Then locally the hypersphere can be treated as a hyperplane and the error is second-order. We have the following relationship when $\epsilon$ is small:
\begin{equation}
  B_{(\frac{1}{\sin \frac{\theta}{2}} + 1)\epsilon \sqrt{d}}(\mathcal{M}_t) \supset B_{\epsilon \sqrt{d}}(\mathcal{M}_{tA}) \cap B_{\epsilon \sqrt{d}}(\mathcal{M}_{tB})
\end{equation}

Then, according to Lemma~\ref{lemma: meanrestriction} and Lemma~\ref{lemma:variancerestriction}:

$\forall \epsilon, \xi, \exists D, \forall d \ge D$, where $\epsilon$ is small, 
\begin{equation}
  \begin{aligned}
    \mathcal{P}\left(d_2\left(\mathbf{x}_t, \mathcal{M}_{tA}\right) < \epsilon \sqrt{d}\right) >& 1 - \frac{\xi}{2} \\
    \mathcal{P}\left(d_2\left(\mathbf{x}_t, \mathcal{M}_{tB}\right) < \epsilon \sqrt{d}\right) >& 1 - \frac{\xi}{2}. \\
  \end{aligned}
\end{equation}
Then,
\begin{equation}
  \begin{aligned}
    \mathcal{P}\left(d_2\left(\mathbf{x}_t, \mathcal{M}_{t}\right) < (\frac{1}{\sin \frac{\theta}{2}} + 1) \epsilon \sqrt{d}\right) \ge& \mathcal{P}\left[d_2\left(\mathbf{x}_t, \mathcal{M}_{tA}\right) < \epsilon \sqrt{d}, d_2\left(\mathbf{x}_t, \mathcal{M}_{tB}\right) < \epsilon \sqrt{d}\right]\\
  =& \mathcal{P}\left(d_2\left(\mathbf{x}_t, \mathcal{M}_{tA}\right) < \epsilon \sqrt{d}\right) + \mathcal{P}\left(d_2\left(\mathbf{x}_t, \mathcal{M}_{tB}\right) < \epsilon \sqrt{d}\right) \\
    &-  \mathcal{P}\left[d_2\left(\mathbf{x}_t, \mathcal{M}_{tA}\right) < \epsilon \sqrt{d} \text{ or } d_2\left(\mathbf{x}_t, \mathcal{M}_{tB}\right) < \epsilon \sqrt{d}\right]\\
    >& 1- \frac{\xi}{2} +  1- \frac{\xi}{2} - 1 \\
    =& 1 - \xi
  \end{aligned}
\end{equation}
\hfill \ensuremath{\Box}

Let $\hat{\alpha}^2_t = \bar{\alpha}_t$ and $\hat{\beta}^2_t = 1 -  \bar{\alpha}_t$ and only consider the block$(c, i, j)$, We can get the Lemma~\ref{lemma:detailed_compact_manifold}.
\subsection{Proof of Lemma~\ref{lemma:separable}}
\textit{Proof.}
We just use the following hyperplane: 
\begin{equation}
  \operatorname{\mu}[\mathbf{x}] = \hat{\alpha}_{\frac{t+t'}{2}} \operatorname{\mu}[\mathbf{y}_0]
\end{equation}

Then, because$ \hat{\alpha}_{t} $ monotonically decreasing with $t$ in VP-SDE. Therefore, The hyperplanes $\operatorname{\mu}[\mathbf{x}] = \hat{\alpha}_{t} \operatorname{\mu}[\mathbf{y}_0]$ and $\operatorname{\mu}[\mathbf{x}] = \hat{\alpha}_{t'} \operatorname{\mu}[\mathbf{y}_0]$ are on different sides of the given hyperplane. As a consequence, $\mathcal{M}_t$ and $\mathcal{M}_{t'}$ are on different sides of the given hyperplane.
\hfill \ensuremath{\Box}

\subsection{Proof of Proposition~\ref{proposition: properties of adjacent map}}
  
\textbf{$\exists ! \mathbf{v_{\mathbf{x}_t}} \in N_{\mathbf{x}_t} \mathcal{M}_t$ that $\mathbf{x}_t + \mathbf{v_{\mathbf{x}_t}} \in \mathcal{M}_{t - 1}$.}

\textit{Proof.} We consider the 2-dim normal space $N_{\mathbf{x}_t} \mathcal{M}_t$. Easy to show that it has these two orthogonal basis vectors, $\left[\mathbf{x}_t - \mu[\mathbf{x}_t](1,1,...,1)\right]$ and $\mu[\mathbf{x}_t](1,1,...,1)$. There are only two points in 2-dim normal space $N_{\mathbf{x}_t} \mathcal{M}_t$ that are in $\mathcal{M}_{t - 1}$ because there are only two points, in 2-dim normal space $N_{\mathbf{x}_t} \mathcal{M}_t$ meet the following conditions:
\begin{itemize}
    \item The distance between this point to  $\mu[\mathbf{x}_{t-1}](1,1,...,1)$ is $C_0$.
    \item The line connecting this point to $\mu[\mathbf{x}_{t-1}](1,1,...,1)$ is perpendicular to $(1,1,...,1)$.
\end{itemize}

And there is only one of them near $\mathbf{x}_t$.

Thus, $\exists ! \mathbf{v_{\mathbf{x}_t}} \in N_{\mathbf{x}_t} \mathcal{M}_t$ that $\mathbf{x}_t + \mathbf{v_{\mathbf{x}_t}} \in \mathcal{M}_{t - 1}$.

\hfill \ensuremath{\Box}

\textbf{$N_{\mathbf{x}_t} \mathcal{M}_t = N_{\mathbf{x}_t + \mathbf{v_{\mathbf{x}_t}}} \mathcal{M}_{t-1}$.}

\textit{Proof.} Because $\mathbf{x}_t + \mathbf{v_{\mathbf{x}_t}} \in N_{\mathbf{x}_t} \mathcal{M}_t$, $\mu[\mathbf{x}_t + \mathbf{v_{\mathbf{x}_t}}] (1,1,...,1) \ in N_{\mathbf{x}_t} \mathcal{M}_t$, and they are two orthogonal basis vectors of $N_{\mathbf{x}_t + \mathbf{v_{\mathbf{x}_t}}} \mathcal{M}_{t-1}$. 

Therefore, $N_{\mathbf{x}_t} \mathcal{M}_t = N_{\mathbf{x}_t + \mathbf{v_{\mathbf{x}_t}}} \mathcal{M}_{t-1}$.

\hfill \ensuremath{\Box}

\textbf{$\mathbf{v_{\mathbf{x}_t}}$ is a transition map from $ T_{\mathbf{x}_t} \mathcal{M}_t$ to $T_{\mathbf{x}_t + \mathbf{v_{\mathbf{x}_t}}} \mathcal{M}_{t-1}$ .}

\textit{Proof.} Because $N_{\mathbf{x}_t} \mathcal{M}_t = N_{\mathbf{x}_t + \mathbf{v_{\mathbf{x}_t}}} \mathcal{M}_{t-1}$, $ T_{\mathbf{x}_t} \mathcal{M}_t$ and $T_{\mathbf{x}_t + \mathbf{v_{\mathbf{x}_t}}} \mathcal{M}_{t-1}$ are parallel, and $\mathbf{v_{\mathbf{x}_t}}$ maps $\mathbf{x}_t$ to $\mathbf{x}_t+\mathbf{v_{\mathbf{x}_t}}$.

Therefore, $\mathbf{v_{\mathbf{x}_t}}$ is a transition map from $ T_{\mathbf{x}_t} \mathcal{M}_t$ to $T_{\mathbf{x}_t + \mathbf{v_{\mathbf{x}_t}}} \mathcal{M}_{t-1}$ .

\hfill \ensuremath{\Box}

\textbf{$\mathbf{v_{\mathbf{x}_t}}$ is determained with $\mathbf{f}(\cdot, \cdot), \mathbf{x_t}, \mathbf{y}_0$ and $g(\cdot)$. }

\textit{Proof.} As proved in~\cite{sarkka2019applied}, the means and covariances of linear SDEs can be transformed to corresponding ODEs. Therefore, suppose $\mathbf{y_t} = \hat{\alpha}_t\mathbf{y}_0 + \hat{\beta}_t \mathbf{z}_t$, $\mathbf{y_{t-1}} = \hat{\alpha}_{t-1}\mathbf{y}_0 + \hat{\beta}_{t-1} \mathbf{z}_{t-1}$, all the coeffients are determained by $\mathbf{f}(\cdot, \cdot), g(\cdot)$.
For clarity, we will represent $\mathbf{v_{\mathbf{x}_t}}$ with $\mathbf{x}_t, \mathbf{y}_0$ and the coefficients above.

In fact, it is easy to show that 
\begin{equation}
    \sqrt{\hat{\alpha}_{t-1}^2Var[\mathbf{y}_0] + \hat{\beta}_{t-1}^2}\left(\frac{\mathbf{x}_t -\hat{\alpha}_t \operatorname{\mu}(\mathbf{y}_0)}{\sqrt{\hat{\alpha}_{t}^2Var[\mathbf{y}_0] + \hat{\beta}_{t}^2}}\right)+\hat{\alpha}_{t-1} \operatorname{\mu}(\mathbf{y}_0)
\end{equation}

is in both $N_{\mathbf{x}_t} \mathcal{M}_t$ and $\mathcal{M}_{t-1}$, and is the one in $N_{\mathbf{x}_t} \mathcal{M}_t \cap \mathcal{M}_{t-1}$ near $\mathbf{x}_t$. Therefore, 
\begin{equation}
    \mathbf{v_{\mathbf{x}_t}} =     \sqrt{\hat{\alpha}_{t-1}^2Var[\mathbf{y}_0] + \hat{\beta}_{t-1}^2}\left(\frac{\mathbf{x}_t -\hat{\alpha}_t \operatorname{\mu}(\mathbf{y}_0)}{\sqrt{\hat{\alpha}_{t}^2Var[\mathbf{y}_0] + \hat{\beta}_{t}^2}}\right)+\hat{\alpha}_{t-1} \operatorname{\mu}(\mathbf{y}_0) - \mathbf{x}_t
\end{equation}

\hfill \ensuremath{\Box}

\subsection{Proof of Remark~\ref{remark, f is in normal space}}

\textit{Proof.} Equivalently, we prove that $\mathbf{x}_t \in N_{\mathbf{x}_t} \mathcal{M}_t$. Consider $\mathcal{M}_{tA}$ and $\mathcal{M}_{tB}$ restricted with one of the equations in Eqn.~\eqref{eqn. for define separated manifolds}. We do the following decomposition of $\mathbf{x}_t$
\begin{equation}
    \begin{aligned}
        \mathbf{x}_t = \left[\mathbf{x}_t - \mu[\mathbf{x}_t](1,1,...,1)\right] + \mu[\mathbf{x}_t](1,1,...,1)
    \end{aligned}
\end{equation}
Easy to know that $\left[\mathbf{x}_t - \mu[\mathbf{x}_t](1,1,...,1)\right] \in N_{\mathbf{x}_t}\mathcal{M}_{tB}$ and $\mu[\mathbf{x}_t](1,1,...,1) \in N_{\mathbf{x}_t}\mathcal{M}_{tA}$. Because $\mathcal{M}_t = \mathcal{M}_{tA} \cap \mathcal{M}_{tB}$, thus the two compoments are all $\in N_{\mathbf{x}_t}\mathcal{M}_{t}$ and then $\mathbf{x}_t \in N_{\mathbf{x}_t} \mathcal{M}_t$ .
\hfill \ensuremath{\Box}

\subsection{Proof of Remark~\ref{remark: expection}}

\textit{Proof.}
\begin{equation}
    \begin{aligned}
  \mathbb{E}[\mathbf{\Phi}(\mathbf{y}_t)] &= \mathbb{E}[\otimes_{c,i,j}\mu[\mathbf{y}_t^{c, i, j}]] \\
  &= \otimes_{c,i,j}\mathbb{E}[\operatorname{\mu}[\mathbf{y}_t^{c, i, j}]]\\
  &= \otimes_{c,i,j}\sqrt{\bar{\alpha}_t}\operatorname{\mu}[\mathbf{y}_0^{c, i, j}]
    \end{aligned}
\end{equation}
\hfill \ensuremath{\Box}

\section{Details about SDDM}
\begin{assumption}

  Suppose $\mathbf{s}(\cdot, \cdot): \mathbb{R}^D \times \mathbb{R} \rightarrow \mathbb{R}^D$ is the score-based model, $\mathbf{f}(\cdot, \cdot): \mathbb{R}^D \times \mathbb{R} \rightarrow \mathbb{R}^D$ is the drift coefficient, $g(\cdot): \mathbb{R} \rightarrow \mathbb{R}$ is the diffusion coefficient, and  $\mathcal{E}(\cdot, \cdot, \cdot): \mathbb{R}^D \times \mathbb{R}^D \times \mathbb{R} \rightarrow \mathbb{R}$ is the energy function. $\mathbf{y}_0$ is the given source image.
  
  Like previous works~\cite{zhao2022egsde,choi2021ilvr, meng2022sdedit}, we define a valid conditional distribution $p\left(\mathbf{x}_0 \mid \mathbf{y}_0\right)$ under following assumptions:
  \begin{itemize}
    \item $\exists C>0, \forall t \in \mathbb{R}, \forall \mathbf{x}, \mathbf{y}_0\in \mathbb{R}^D:\|f(\mathbf{x}, t)-f(\mathbf{y}_0, t)\|_2 \leq C\|\mathbf{x}-\mathbf{y}\|_2$.
    \item $\exists C>0, \forall t, s \in \mathbb{R}, \forall \mathbf{x} \in \mathbb{R}^D:\|f(\mathbf{x}, t)-f(\mathbf{x}, s)\|_2 \leq C|t-s|$.
    \item $\exists C>0, \forall t \in \mathbb{R}, \forall \mathbf{x}, \mathbf{y}_0\in \mathbb{R}^D:\|\mathbf{s}(\mathbf{x}, t)-\mathbf{s}(\mathbf{y}_0, t)\|_2 \leq C\|\mathbf{x}-\mathbf{y}\|_2$.
    \item $\exists C>0, \forall t, s \in \mathbb{R}, \forall \mathbf{x} \in \mathbb{R}^D:\|\mathbf{s}(\mathbf{x}, t)-\mathbf{s}(\mathbf{x}, s)\|_2 \leq C|t-s|$.
    \item $\exists C>0, \forall t \in \mathbb{R}, \forall \mathbf{x}, \mathbf{y}_0\in \mathbb{R}^D:\left\|\nabla_{\mathbf{x}} \mathcal{E}\left(\mathbf{x}, \mathbf{y}_0, t\right)-\nabla_{\mathbf{y}} \mathcal{E}\left(\mathbf{y}_0, \mathbf{y}_0, t\right)\right\|_2 \leq C\|\mathbf{x}-\mathbf{y}\|_2$.
    \item $\exists C>0, \forall t, s \in \mathbb{R}, \forall \mathbf{x} \in \mathbb{R}^D:\left\|\nabla_{\mathbf{x}} \mathcal{E}\left(\mathbf{x}, \mathbf{y}_0, t\right)-\nabla_{\mathbf{x}} \mathcal{E}\left(\mathbf{x}, \mathbf{y}_0, s\right)\right\|_2 \leq C|t-s|$.
    \item $\exists C>0, \forall t, s \in \mathbb{R}:|g(t)-g(s)| \leq C|t-s|$.
  \end{itemize}
  
  \end{assumption}

\subsection{Details about Pre-Trained Diffusion Models}
We use two pre-trained diffusion models and a VGG model.

In the Cat $\rightarrow$ Dog task and Wild $\rightarrow$ Dog task, we use the public pre-trained model provided in the official code \href{https://github.com/jychoi118/ilvr\_adm}{https://github.com/jychoi118/ilvr\_adm} of ILVR~\cite{choi2021ilvr}.

In the Male $\rightarrow$ Female task, we use the public pre-trained model provided in the official code \href{https://github.com/ML-GSAI/EGSDE}{https://github.com/ML-GSAI/EGSDE} of EGSDE~\cite{zhao2022egsde}.

Our energy function uses the pre-trained VGG net provided in the unofficial open source code \href{https://github.com/naoto0804/pytorch-AdaIN}{https://github.com/naoto0804/pytorch-AdaIN} of AdaIN~\cite{huang2017arbitrary}.

\subsection{Details about Our Default Model Settings}
\label{section: details about settings}
Our default SDDM settings:
\begin{itemize}
    \item Using BAdaIN to construct manifolds.
    \item Using multi-optimization on manifolds.
    \item $\lambda = 2, \lambda_1 = 25$.
    \item Using $\epsilon$ Policy 3.
    \item Blocks are $16\times 16$.
    \item $T_0 = 0.5 T$.
    \item 100 diffusion steps.
\end{itemize}

SDDM$^{\dagger}$ sets $T_0 = 0.6 T$.

\subsection{Implementation Details about Solving Problem~\eqref{equ: optimization target}}

To simplify the process, we denote all the vectors as $\{\mathbf{v}_i\}$, and coefficients as $\{\lambda_i\}$, and rewrite Problem~\eqref{equ: optimization target} as 

\begin{equation}
 \min _{\substack{\lambda_1, \lambda_2, \ldots, \lambda_n \ge 0 \\\lambda_1 +  \lambda_2 + \ldots + \lambda_n = 1} }\left\{\left\| \sum_{i=1}^n \lambda_i \mathbf{v}_i\right\|_2^2\right\}.
 \end{equation}

When there are only two vectors (in our situation) and no restriction $\lambda_1, \lambda_2, \ldots, \lambda_n \ge 0$, we can get the following analytical solution:
\begin{equation}
    \hat\lambda_1^* = \frac{(\mathbf{v}_2 - \mathbf{v}_1)^T\mathbf{v}_2}{\|\mathbf{v}_2 - \mathbf{v}_1\|_2^2}.
\end{equation}

Therefore, easy to prove that when there are only two vectors, the analytical solution is:
 \begin{equation}
    \lambda_1^* = \min\left(1, \max\left(\frac{(\mathbf{v}_2 - \mathbf{v}_1)^T\mathbf{v}_2}{\|\mathbf{v}_2 - \mathbf{v}_1\|_2^2}, 0\right)\right).
\end{equation}

For general situations, we can apply Frank–Wolfe algorithm on this problem as in~\cite{sener2018multi}.

\section{Details about FID calculation}
\label{sec: fid cal}
The FID is calculated between 500 generated images and the target validation dataset containing 500 images in the Cat $\rightarrow$ Dog and Wild $\rightarrow$ Dog task. The number is 1000 in the Male $\rightarrow$ Female task. All experiments are repeated 5 times to eliminate the randomness.

\section{FID on the Male$\rightarrow$Female task}

It is true that EGSDE with sufficient diffusion steps outperforms our SDDM on the Male$\rightarrow$Female task, it is important to note that the energy function used in EGSDE is strongly pretrained on related datasets and contains significant domain-specific information. In contrast, to demonstrate the effectiveness and versatility of our framework, we intentionally chose to use a weak energy function consisting of only one layer of convolution without any further pretraining. After incorporating the strong guidance function from EGSDE, our method outperforms EGSDE in the FID score, as shown in the following table.

%---------------------------Table-------------------------------------------------------------
\begin{table}[htbp]
\caption{The FID comparison between EGSDE and our SDDM with the same energy guidance function on the Male$\rightarrow$Female task.} 
\label{tab: compare sota fid}
\setlength\tabcolsep{14pt}
\vskip 0.05in
\begin{center}
\begin{small}
\begin{sc}
\begin{tabular}{lcc}
\toprule
Model & FID$\downarrow$\\
\midrule
% SDDM(low-pass filter) &74.37&56.95 &  17.77 & 0.363\\
% SDDM(low-pass filter) &67.79& \textbf{51.49} & \textbf{18.81} & \textbf{0.420}\\
EGSDE & 41.93 ± 0.11\\
SDDM(Ours) &\textbf{40.08 ± 0.13} \\
\bottomrule
\end{tabular}
\end{sc}
\end{small}
\end{center}
\vskip -0.1in
\end{table}
%----------------------------------------------------------------------------------------------

\section{Samples}
\label{section: visual comparisons}
\begin{figure*}[htbp] %H为当前位置，!htb为忽略美学标准，htbp为浮动图形
 \centering %图片居中
 \includegraphics[width=0.7\textwidth]{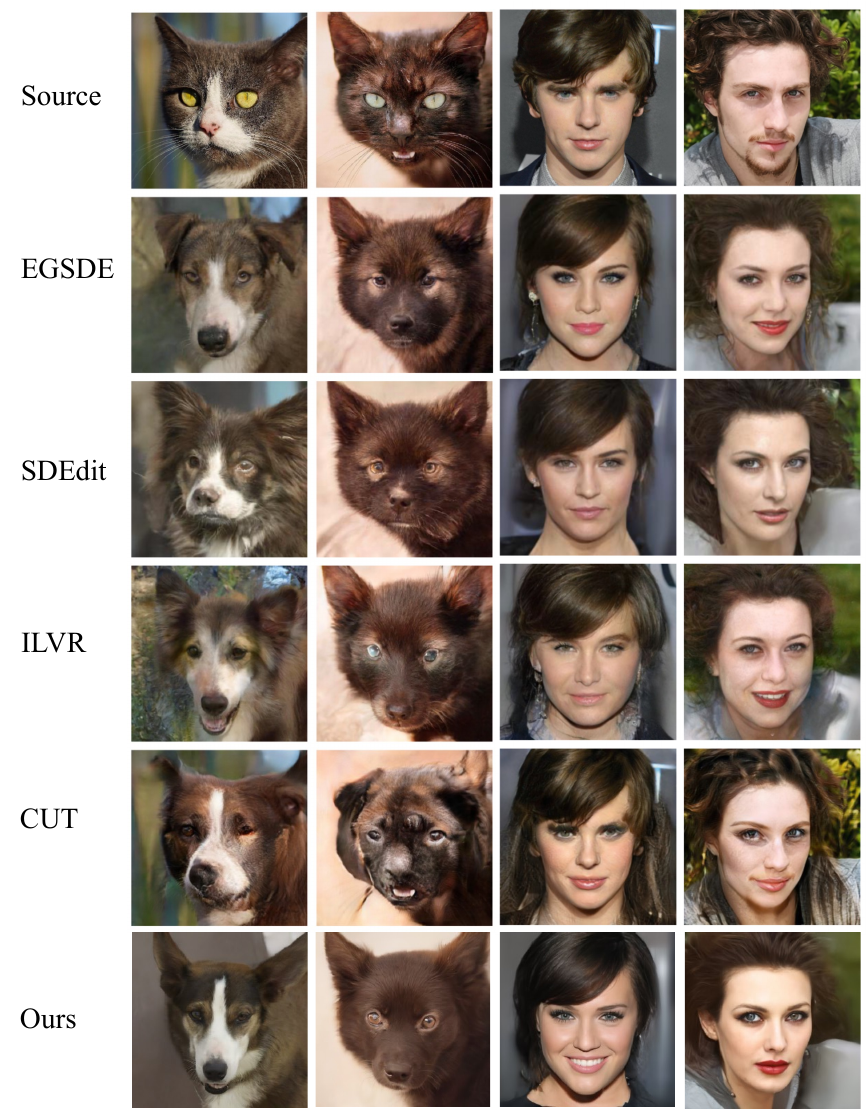} %插入图片，[]中设置图片大小，{}中是图片文件名
 \caption{The visual comparison of different methods.}
 \end{figure*}

\begin{figure*}[htbp] %H为当前位置，!htb为忽略美学标准，htbp为浮动图形
 \centering %图片居中
 \includegraphics[width=0.6\textwidth]{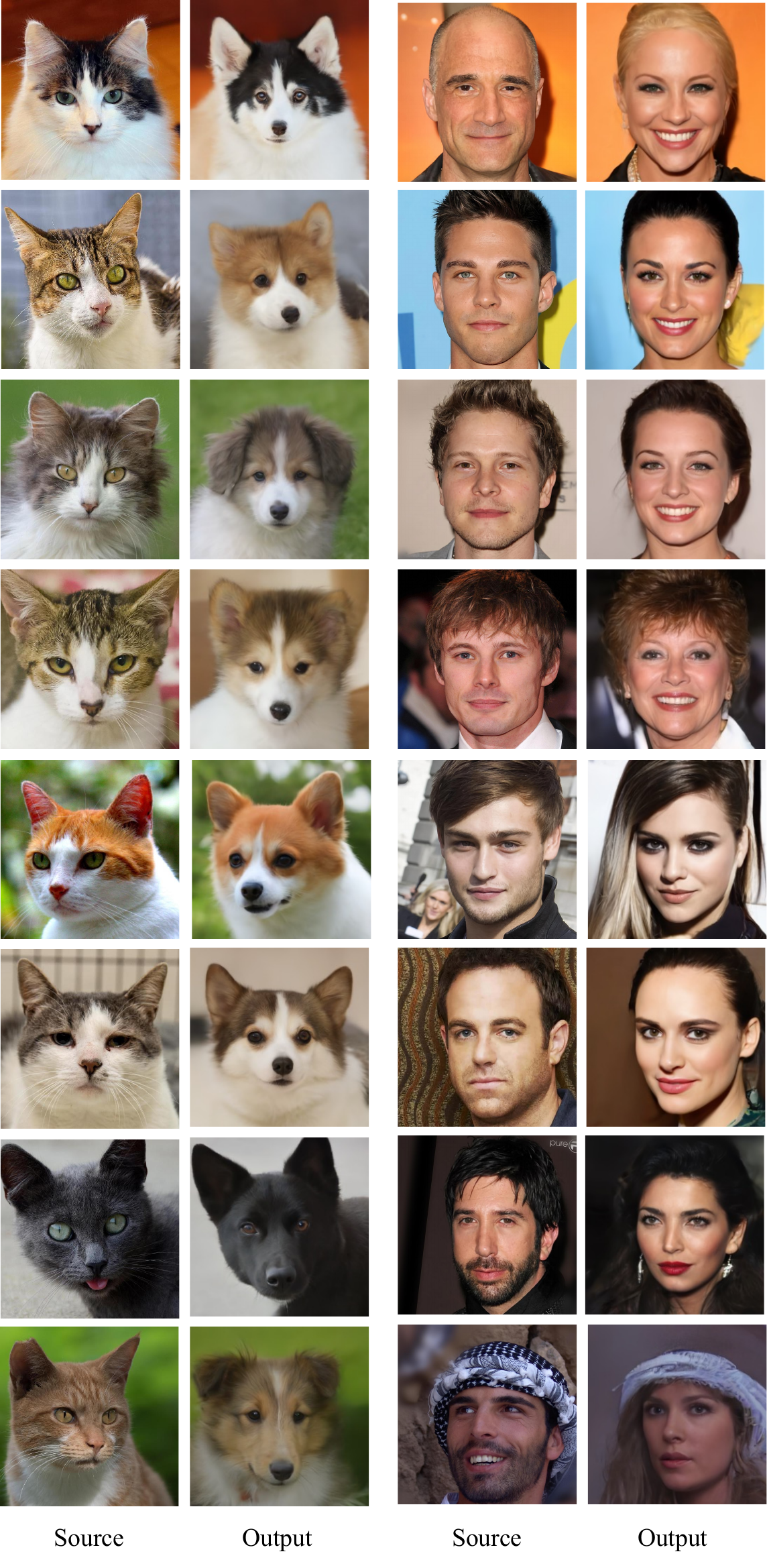} %插入图片，[]中设置图片大小，{}中是图片文件名
 \caption{More random samples of our methods.}
 \end{figure*}

% You can have as much text here as you want. The main body must be at most $8$ pages long.
% For the final version, one more page can be added.
% If you want, you can use an appendix like this one, even using the one-column format.
%%%%%%%%%%%%%%%%%%%%%%%%%%%%%%%%%%%%%%%%%%%%%%%%%%%%%%%%%%%%%%%%%%%%%%%%%%%%%%%
%%%%%%%%%%%%%%%%%%%%%%%%%%%%%%%%%%%%%%%%%%%%%%%%%%%%%%%%%%%%%%%%%%%%%%%%%%%%%%%

\end{document}